\documentclass[journal]{IEEEtran}

\IEEEoverridecommandlockouts                              

\usepackage{amsmath,mathtools}
\usepackage{amsfonts}
\usepackage{amssymb}
\usepackage{balance}
\usepackage{graphicx}
\usepackage[skip=2pt,font=small]{caption}
\usepackage{subcaption}
\usepackage[numbers, square, comma,compress]{natbib}
\usepackage{siunitx}
\usepackage[table]{xcolor} 
\usepackage{color}
\usepackage{footmisc}
\usepackage{dblfloatfix}
\usepackage{url}
\usepackage{nth}
\usepackage{animate}
\usepackage{comment}
\usepackage{tikz}
\usepackage{mathtools} 
\usepackage{lipsum}
\usepackage{multirow}
\usepackage{makecell}
\usepackage{hyperref}
\usetikzlibrary{arrows}
\usepackage[normalem]{ulem}
\usepackage{paralist}
\usepackage{pifont}
\usepackage[ruled,vlined]{algorithm2e}
\usepackage{booktabs, multirow} 
\usepackage{soul}
\usepackage{pgfplots}
\pgfplotsset{compat=1.17}
\newcommand{\rom}[1]{(\expandafter{\romannumeral #1\relax})}

\definecolor{royalazure}{rgb}{0.0, 0.22, 0.66}
\definecolor{mayablue}{rgb}{0.45, 0.76, 0.98}
\DeclareMathOperator*{\argmin}{argmin}

\newcommand{\add}[0]{\textcolor{black}}

\definecolor{somegray}{rgb}{0.5, 0.5, 0.5}
\newcommand{\darkgrayed}[1]{\textcolor{somegray}{#1}}
\makeatletter
\newcommand*\titleheader[1]{\gdef\@titleheader{#1}}
\AtBeginDocument{%
  \let\st@red@title\@title
  \def\@title{%
    \vskip-2em
    \bgroup\normalfont\large\centering\@titleheader\par\egroup
    \vskip1.5em\st@red@title}
}
\makeatother

\titleheader{\darkgrayed{This paper has been accepted for publication at the\\
IEEE Transactions on Robotics, 2022.
\copyright IEEE}}

\title{A Comparative Study of Nonlinear MPC and Differential-Flatness-Based Control for Quadrotor Agile Flight}

\author{Sihao Sun, Angel Romero, Philipp Foehn, Elia Kaufmann and Davide Scaramuzza
\thanks{All authors are with the Robotics and Perception Group, University of Zurich, Switzerland (\protect\url{http://rpg.ifi.uzh.ch}). 
This work was supported by in part by the National Centre of Competence in Research (NCCR) Robotics, through the Swiss National Science Foundation (SNSF), in part by the European Union’s Horizon 2020 Research and Innovation Programme under grant agreement No. 871479 (AERIAL-CORE), and in part by the European Research Council (ERC) under grant agreement No. 864042 (AGILEFLIGHT).
    }
}
\begin{document}
\maketitle

\begin{abstract}
Accurate trajectory-tracking control for quadrotors is essential for safe navigation in cluttered environments.
However, this is challenging in agile flights due to nonlinear dynamics, complex aerodynamic effects, and actuation constraints. 
In this article, we empirically compare two state-of-the-art control frameworks: the nonlinear-model-predictive controller (NMPC) and the differential-flatness-based controller (DFBC), by tracking a wide variety of agile trajectories at speeds up to \add{20~\si{m/s} (i.e., 72~\si{km/h})}. The comparisons are performed in both simulation and real-world environments to systematically evaluate both methods from the aspect of tracking accuracy, robustness, and computational efficiency. 
We show the superiority of NMPC in tracking dynamically infeasible trajectories, at the cost of higher computation time and risk of numerical convergence issues.
For both methods, we also quantitatively study the effect of adding an inner-loop controller using the incremental nonlinear dynamic inversion (INDI) method, and the effect of adding an aerodynamic drag model.
Our real-world experiments, performed in one of the world's largest motion capture systems, demonstrate more than 78\% tracking error reduction of both NMPC and DFBC, indicating the necessity of using an inner-loop controller and aerodynamic drag model for agile trajectory tracking.
\end{abstract}

\section*{Multimedia Material}
The experimental results can be viewed in this video:\\
\url{https://youtu.be/XpuRpKHp_Bk}
\section{Introduction}
\subsection{Motivation}
\begin{figure}[!t]
    \centering
       \begin{subfigure}[b]{0.5\textwidth}
   \includegraphics[width=0.98\textwidth]{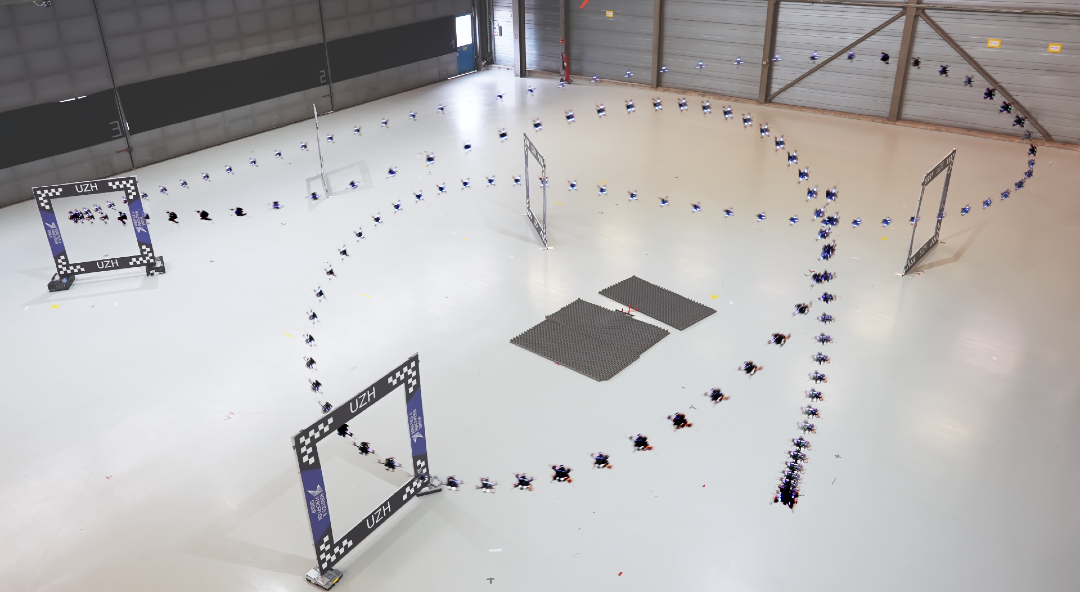}
\end{subfigure}
   \begin{subfigure}[b]{0.5\textwidth}
    \includegraphics[width=0.98\textwidth]{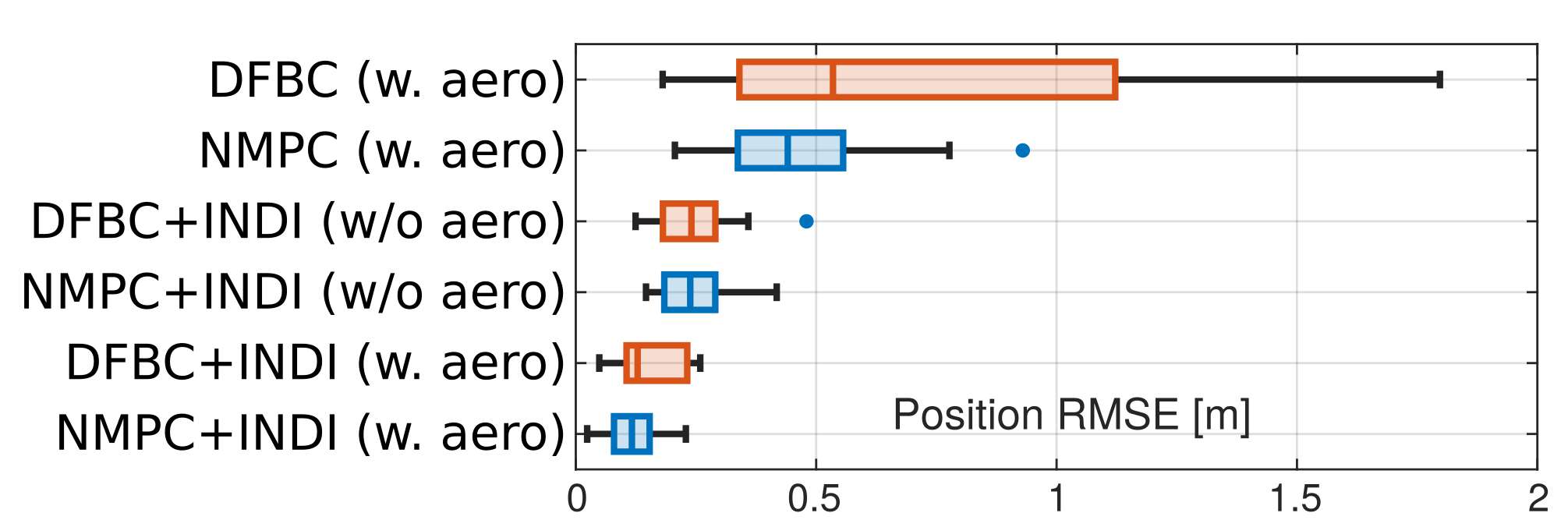}
    \end{subfigure}
    \caption{\textbf{Top:} Our quadrotor tracking a race trajectory. \textbf{Bottom:} Box plot comparing the position tracking root-mean-square-error (RMSE) of NMPC, DFBC, and their variations with INDI inner-loop, \add{with or without considering aerodynamic drag effects}. For each method, data are collected from real-world flights tracking different reference trajectories with speeds up to \add{\textbf{20~\si{m/s}}} (i.e., 72~\si{km/h}) and accelerations up to \textbf{5}g. 
}
    \label{fig:eye_catcher}
\end{figure}

Quadrotors are extremely agile. Exploiting their agility is crucial for time-critical missions, such as search and rescue, monitoring, exploration, aerial delivery,  drone racing, reconnaissance, and even flying cars~\cite{Loianno2020jfr,rajendran2020air}. An accurate trajectory-tracking controller is required to safely execute high-speed trajectories in cluttered environments. However, most approaches struggle to handle joint effects in agile flights, such as nonlinear dynamics, aerodynamic effects, and actuation limits.

Recently, nonlinear model predictive control (NMPC) has drawn much attention for quadrotor control thanks to advances in hardware and algorithmic efficiency~\cite{bicego2020nonlinear,foehn2021time,torrente2021data,kamel2015fast,kamel2017model,murilo2019unified,small2019aerial,nguyen2021model,romero2022mpcc}. NMPC particularly excels in handling control limits, and its predictive nature is believed to be beneficial for trajectory tracking at high speeds~\cite{bicego2020nonlinear}. A recent study has demonstrated its performance in tracking aggressive trajectories up to 20~\si{m/s}~\cite{foehn2021time}.

However, NMPC is computationally extremely demanding compared to the state-of-the-art non-predictive method: the differential-flatness-based controller (DFBC)~\cite{faessler2017differential,tal2020accurate}. 
This method also shows great potential in accurately tracking agile trajectories autonomously.
A recent study has used DFBC to track trajectories up to 12.9~\si{m/s} with 2.1~g accelerations, with only 6.6 centimeters position tracking error~\cite{tal2020accurate}.
Although the state-of-the-art DFBC has not achieved agile flights as fast as NMPC showed in~\cite{foehn2021time}, its high computational efficiency and tracking accuracy render the necessity of applying NMPC for agile trajectory tracking questionable.
Therefore, a comparative study of NMPC and DFBC is needed to provide insights for future research on fully autonomous agile flights, in order to further improve their efficiency and reliability.

\subsection{Contribution}
This article presents the first comparative study of the two state-of-the-art tracking controllers: NMPC and DFBC method, in fast trajectory tracking with speed up to 20~\si{m/s} (i.e., 72~\si{km/h}) and 5~g accelerations.
Specifically, we select the NMPC method that uses the full nonlinear dynamics with proper actuator bounds and regards single rotor thrusts as control inputs. This NMPC has been applied in previous works, such as~\cite{bicego2020nonlinear,torrente2021data}.
For a fair comparison, we improve the DFBC method used in~\cite{faessler2017differential, tal2020accurate} with a control allocation approach using constrained quadratic programming~\cite{johansen2013control} also to consider control input limits.

All experiments are conducted in both simulations and the real world. 
The simulations compare in both ideal and practical conditions with model-mismatch, estimation latency, and external disturbances. 
The real-world experiments are conducted in one of the world's largest motion capture systems, with 30$\times$30$\times$8 m$^3$ flight volume. Multiple agile trajectories are selected as reference, including \add{not only dynamically \textit{feasible} trajectories, but also dynamically  \textit{infeasible} trajectories that require thrusts exceeding the maximum capacity of the quadrotor motors, which is likely to happen in high-speed flights due to model mismatch.} These tests investigate the performance of both methods in the presence of significant high-speed-induced aerodynamic effects, inevitable system latency, and on an onboard embedded computer. 

The experimental results reveal that NMPC is considerably more computationally demanding, and more prone to suffer from numerical convergence issues in the presence of large external force disturbances. However, NMPC also excels in tracking dynamically \textit{infeasible} trajectories, making it more suitable for tracking time-optimal trajectories that violate the rotor thrust constraints.

In addition, this study also highlights the importance of implementing an inner-loop controller for robustification. Specifically, we select the incremental nonlinear dynamic inversion (INDI) method as the inner-loop angular controller, thanks to its simplicity in implementation and demonstrated robustness in various real-world experiments~\cite{smeur2016adaptive,sun2020incremental} including a combination with DFBC~\cite{tal2020accurate}. As for NMPC, differently from the state-of-the-art using a PID as the low-level control~\cite{foehn2021time}, we propose a method to hybridize NMPC with INDI that considers the real input limits of the quadrotor instead of constraints on virtual inputs. 
Real-world flight results demonstrate more than 78\% position tracking error reduction of NMPC and DFBC with an INDI inner-loop.
We also reveal that a well-selected inner-loop controller is more crucial than simply considering the aerodynamic effects, as is compared in Fig.~\ref{fig:eye_catcher}.

Apart from the technical contribution, this paper can also be regarded as a tutorial for non-expert readers on agile quadrotor flight. We encourage the practitioners to use the presented results as a baseline for further development of both DFBC and NMPC approaches.

\section{Related Work}
In this paper, we classify the trajectory tracking controller into two categories: non-predictive and predictive methods. 
While the predictive methods encode multiple future time-steps into the control command, the non-predictive methods only track a single reference step. 
In the following, we review works towards improving quadrotor trajectory tracking accuracy from these two different categories. 
A more comprehensive survey of quadrotor position and attitude control methods can be found in the literature (e.g., see~\cite{nascimento2019position, lee2017trajectory}). 

\subsection{Non-predictive Quadrotor Trajectory Tracking Control}
Unlike most fixed-wing aircraft, quadrotors are inherently unstable. Therefore, the initial work of quadrotor control aimed at achieving stable hovering and near-hover flights. 
Thanks to the small-angle assumptions in these conditions, linear control methods such as PID and LQR demonstrate sufficiently good performance (see, e.g.,~\cite{khatoon2014pid, dong2013modeling}).

However, as the requirements for agile flight emerges, these assumptions are no longer valid. 
Nonlinearities from the attitude dynamics are the first problem to tackle. For this reason, nonlinear flight controllers are proposed, such as feedback linearization~\cite{voos2009nonlinear} and backstepping~\cite{madani2006backstepping}. 
In order to cope with the singularities of Euler angles as the nonlinear attitude representation, quaternions are widely adopted to parametrize the attitude~\cite{fresk2013full}. 
In addition, the authors of \cite{lee2010geometric} propose the geometric tracking controller to directly control the quadrotor on the manifold of the special Euclidean group SE(3), showing almost globally asymptotic tracking of the position, velocity, and attitude of the quadrotor.

A seminal work~\cite{mellinger2011minimum} reveals that quadrotors are differentially flat systems.
By virtue of this property, given the time-parameterized 3D path, one can derive the reference attitude, angular rate, and accelerations. 
These references can be sent to a closed-loop flight controller as feedforward terms, while additional feedback control is required to address model mismatch and external disturbances. 
As such, differential-flatness based controller (DFBC) has significantly improved the tracking performance at relatively high speeds~\cite{faessler2017differential,tal2020accurate}. 

As the flying speed increases, a quadrotor starts experiencing non-negligible aerodynamic effects, including drag~\cite{mahony2012multirotor}, aerodynamic torque~\cite{sun2019quadrotor}, and variation of thrusts~\cite{huang2009aerodynamics}. 
Authors of \cite{faessler2017differential} show that the aerodynamic drag does not affect the differential flatness of a quadrotor. 
Thus they adopt a first order aerodynamic model with feedforward terms derived from the reference trajectory to improve the tracking performance. 
Since an accelerometer can directly read external forces, \cite{tal2020accurate} leverages accelerometer measurements to improve the tracking accuracy instead of resorting to an aerodynamic drag model. This method also demonstrates remarkable performance in disturbance rejection and platform adaptability.

Effectively handling control input limits is a remaining challenge for non-predictive methods, including DFBC. So far, existing methods have prioritized the position tracking over heading using various approaches, such as redistributed pseudo inversion ~\cite{faessler2016thrust}, weighted-least-square allocation~\cite{smeur2017prioritized}, control-prioritization method~\cite{brescianini2018tilt,tal2020accurate}, and constrained-quadratic-programming allocation~\cite{zaki2017trajectory}. While these methods can mitigate the actuator saturation effect when the trajectories are dynamically \textit{feasible}, its performance in tracking dynamically \textit{infeasible} trajectories is still questionable. In this work, we will push the limit of DFBC to conduct agile trajectory tracking tasks to a much higher flight speed and acceleration than the state-of-the-art, and study its performance in tracking dynamically \textit{infeasible} trajectories where violating rotor thrust limits becomes inevitable.

\subsection{Model Predictive Control for Quadrotor Trajectory Tracking}
Model predictive control (MPC) is a prevalent method in robots thanks to its predictive nature and ability to handle input constraints\cite{nguyen2021model,kamel2017linear}. 
It generates control commands in receding horizon fashion, which minimizes the tracking error in the predicted time horizon by solving constrained optimization problems.

However, MPC is very computationally demanding compared with non-predictive methods. Especially, using nonlinear-MPC (NMPC) with a full-state nonlinear model of a quadrotor was computationally intractable onboard early-age unmanned-aerial vehicles (UAVs). 
For this reason, linear-MPC (LMPC) is adopted in many studies for only position control~\cite{bangura2014real}, or control linearized model based on small-angle assumptions~\cite{alexis2014trajectory}. 
Therefore, these LMPC methods cannot fully capture nonlinearities in rotational dynamics, and underperforms NMPC methods~\cite{nguyen2021model,kamel2017linear}

Separating flight controllers into high-level position control and low-level attitude control is another common approach to simplify the model and alleviate the computational load of NMPC~\cite{kamel2015fast,kamel2017model,small2019aerial}. 
However, in such a cascaded control architecture, the high-level controller cannot precisely capture the real quadrotor capability because they often ignore the effects of system limitation~(such as \cite{kamel2015fast}), or introduce a virtual constraint on states~\cite{small2019aerial}. 
Consequently, the command fed into the lower-level controller is either too conservative to achieve agile flights, or over-aggressive that causes instability.

Thanks to the recent development in hardware and nonlinear optimization solvers~\cite{houska2011acado,chen2019matmpc,andersson2019casadi}, running NMPC with a nonlinear full dynamic model becomes computationally tractable on an embedded computer.
Hence, recently, some studies start using the full-nonlinear dynamic model, and regarding single rotor thrusts as inputs for NMPC, thus are able to fully exploit quadrotors capability~\cite{bicego2020nonlinear,foehn2021time, torrente2021data}. 
These methods either directly use the optimized single rotor thrust commands~\cite{bicego2020nonlinear}, or send intermediate states from the solution (such as the angular rates) to a low-level controller~\cite{foehn2021time,torrente2021data}.
A recent study~\cite{foehn2021time} demonstrates the ability of the full-model NMPC with a PID low-level controller in tracking a pre-planned race trajectory at speed up 20~\si{m/s} which surpasses the top speed of 12.9~\si{m/s} reported in \cite{tal2020accurate} using DFBC, in spite of a much larger tracking error (0.7~\si{m} v.s. 0.066~\si{m}).

Although running NMPC is realizable on modern embedded computers, it still requires significantly more computational resources than non-predictive methods represented by DFBC. 
For this reason, NMPC may suffer from numerical convergence issues, especially when the platform lacks a sufficient computational budget.
Moreover, the advantage of NMPC becomes questionable as DFBC can also address input limits using the control allocation technique and generate feedforward control leveraging differential-flatness property.
Therefore, it is necessary to compare these two methodologies and understand at what conditions each approach is preferable to provide insights and recommendations for future applications.

\section{Preliminaries}
\label{sec:preliminaries}
\subsection{Notations}
\begin{figure}
    \centering
    \includegraphics[width=0.4\textwidth]{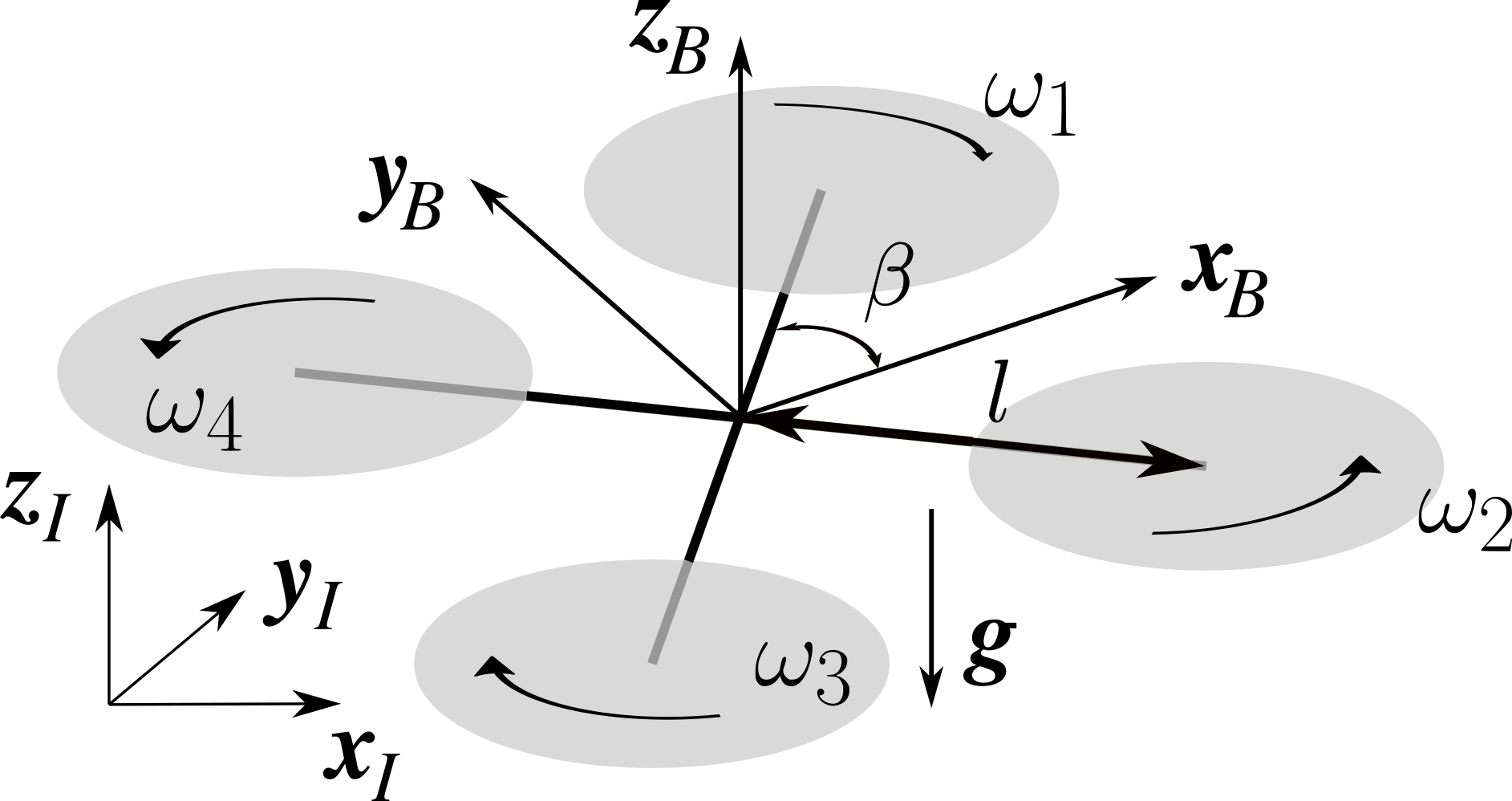}
    \caption{Coordinate definitions and propeller numbering convention.}
    \label{fig:drone}
\end{figure}

Throughout the paper, we use subscription $r$ to indicate the reference variables derived from the reference trajectory.
Subscription $d$ indicates the desired value, which is calculated from a higher-level controller. We use bold lowercase letters to represent vectors and bold uppercase letters for matrices; otherwise, they are scalars. 
Two right-handed coordinate frames are used in this paper: they are the inertial-frame $\mathcal{F}_I:$ $\{\boldsymbol{x}_I,\boldsymbol{y}_I,\boldsymbol{z}_I\}$ with $\boldsymbol{z}_I$ pointing upward opposite to the gravity, and the body-frame $\mathcal{F}_B:\{\boldsymbol{x}_B, \boldsymbol{y}_B, \boldsymbol{z}_B\}$ with $\boldsymbol{x}_B$ pointing forward and $\boldsymbol{z}_B$ aligned with the collective thrust direction (see Fig.~\ref{fig:drone}). 
Vectors with superscription $B$ are expressed in the body-frame; others without any superscription are expressed in the inertial-frame. The rotation from $\mathcal{F}_I$ to $\mathcal{F}_B$ is represented by rotational matrix $\boldsymbol{R}(\boldsymbol{q})=[\boldsymbol{x}_B,\boldsymbol{y}_B,\boldsymbol{z}_B] \in \mathrm{SO}(3)$ parameterized by quaternion $\boldsymbol{q}= [q_w,~q_x,~q_y,~q_z]^T \in \mathbb{S}^3$. 
\add{We use subscript $\{x,y,z\}$ to represent the imaginary components of a quaternion, namely $\boldsymbol{q}_{x,y,z} = [q_x,~q_y,~q_z]^T$.} 
Operator $\mathrm{diag}(A_1,A_2,...,A_n)$ denotes a diagonal matrix with scalars or matrices ($A_1,A_2,...,A_n$) as diagonal entries. 
\subsection{Quadrotor Model}
\subsubsection{Quadrotor Rigid-Body Model}
The quadrotor model is established using 6-DoF rigid body kinematic and dynamic equations. For translational dynamics, we have
\begin{equation}
    \ddot{\boldsymbol{\xi}} =  (T\boldsymbol{z}_B + \boldsymbol{f}_a) / m + \boldsymbol{g},
    \label{eq:translate}
\end{equation}
where $\boldsymbol{\xi}$ denotes the position of the quadrotor center of gravity (CoG); $T$ and $m$ are the collective thrust and total mass respectively; 
$\boldsymbol{g}\in$ is the gravitational vector; $\boldsymbol{f}_a$ indicates the exogenous aerodynamic drag force during high-speed flights.

The rotational kinematic and dynamic equations are expressed as
\begin{equation}
\dot{\boldsymbol{q}} = 
\frac{1}{2}\boldsymbol{q} \otimes
\left[\begin{array}{c}
     0  \\
     \boldsymbol{\Omega}^B 
\end{array}\right],
\label{eq:rot_kine}
\end{equation}
\add{
\begin{equation}
\boldsymbol{I}_v\dot{\boldsymbol{\Omega}}^B = \boldsymbol{I}_v\boldsymbol{\alpha}^B= -\boldsymbol{\Omega}^B \times \boldsymbol{I}_v\boldsymbol{\Omega}^B + \boldsymbol{\tau} + \boldsymbol{d}_\tau,
\label{eq:rot_dyn}
\end{equation}}where $\otimes$ denotes the quaternion multiplication operator.
$\boldsymbol{\Omega}$ is the angular velocity of $\mathcal{F}_B$ with respect to $\mathcal{F}_I$. Its derivative, namely angular acceleration, is denoted by $\boldsymbol{\alpha}$ . Throughout the paper, we use its projection on $\mathcal{F}_B$, namely $\boldsymbol{\Omega}^B = [\Omega_x,~\Omega_y,~\Omega_z]^T$, since its directly measurable from the inertial measurement unit (IMU).
$\boldsymbol{I}_v$ indicates the inertia matrix of the entire quadrotor. 
$\boldsymbol{\tau}$ is the resultant torque generated by rotors. \add{$\boldsymbol{d}_\tau$ is the model uncertainties on the body torque, which can come from high-order aerodynamic effects, center of gravity bias, or distinction among rotors.}

The collective thrust and rotor generated torques are functions of rotor speeds:
\begin{equation}
\left[
\begin{array}{c}
     T  \\
     \boldsymbol{\tau} 
\end{array}\right] = \boldsymbol{G}_1 \boldsymbol{u} + \boldsymbol{G}_2\dot{\boldsymbol{\omega}} + \boldsymbol{G}_3(\boldsymbol{\Omega}){\boldsymbol{\omega}} 
\label{eq:equation_T_tau}
\end{equation}
where 
\begin{equation}
\boldsymbol{u} = c_t \boldsymbol{\omega}^{\circ 2}
\label{eq:equation_u}
\end{equation}
represents the thrust generated by each rotor and $^{\circ}$ indicates the Hadamard power. \add{$c_t$ is the thrust coefficient.}
$\boldsymbol{\omega}$ is the vector of angular rates of each propeller. $\boldsymbol{G}_1$ to $\boldsymbol{G}_3$ are matrices defined as 
\begin{equation}
\boldsymbol{G_1} = \left[ 
\begin{array}{cccc} 
    1 & 1 & 1 & 1 \\
    l \sin{\beta} & -l \sin{\beta} & -l \sin{\beta} &  l \sin{\beta}\\
    -l \cos{\beta} & -l \cos{\beta} & l \cos{\beta} &  l \cos{\beta}\\
    c_q/c_t & -c_q/c_t & c_q/c_t & -c_q/c_t
\end{array} \right]
\end{equation}
\begin{equation}
\boldsymbol{G_2} = \left[\begin{array}{cccc}
    0 & 0 & 0 & 0 \\
    0 & 0 & 0 & 0 \\
    0 & 0 & 0 & 0 \\
    I_p & -I_p & I_p & -I_p \\
\end{array} 
\label{eq:G2}
\right]
\end{equation}
\begin{equation}
\boldsymbol{G_3} = \left[\begin{array}{cccc}
    0 & 0 & 0 & 0 \\
    I_p\Omega_y & -I_p\Omega_y & I_p\Omega_y & -I_p\Omega_y \\
    -I_p\Omega_x & I_p\Omega_x & -I_p\Omega_x & I_p\Omega_x \\
    0 & 0 & 0 & 0 \\
\end{array} 
\label{eq:G3}
\right]
\end{equation}
where $c_q$ is the torque coefficient; $\beta$ and $l$ are geometric parameters defined as per Fig.~\ref{fig:drone}. $I_p$ is the inertia of the rotor around $\boldsymbol{z}_B$. 

Terms $\boldsymbol{G}_2\dot{\boldsymbol{\omega}}$ and $\boldsymbol{G}_3(\boldsymbol{\Omega}) \boldsymbol{\omega}$ are torques respectively due to angular acceleration of rotors and gyroscopic effects, which are usually neglected for controller design. However, in Sec~\ref{sec:methodologies_indi}, we will revisit the INDI method~\cite{tal2020accurate} for angular acceleration control that takes into account effects of the inertial torque term $\boldsymbol{G}_2\dot{\boldsymbol{\omega}}$.

\subsubsection{Aerodynamic Drag Model}
Quadrotors during high-speed flight experience significant aerodynamic drag forces, which need to be precisely modeled to improve tracking accuracy while minimizing the computational overhead.
In this work, we use an aerodynamic drag model which captures the major effects, and is proved effective in works such as \cite{faessler2017differential}.
\begin{equation}
\boldsymbol{f}_a^B =\left[\begin{array}{c}
     -k_{d,x} v_x  \\
     -k_{d,y} v_y  \\
     -k_{d,z} v_z + k_{h}\left(v_x^2 + v_y^2\right)
\end{array}\right]
\label{eq:aero_force}
\end{equation}
where $[v_x,~v_y,~v_z] = \boldsymbol{R}(\boldsymbol{q})^T \dot{\boldsymbol{\xi}}$ (i.e., the projection of velocity in the body frame; here we assume zero wind-speed). $k_{d,x,y,z}$ and $k_h$ are positive parameters can be identified from flight data.

\section{Methodologies}
\label{sec:methodologies}
\begin{figure*}
    \centering
    \includegraphics[scale=0.8]{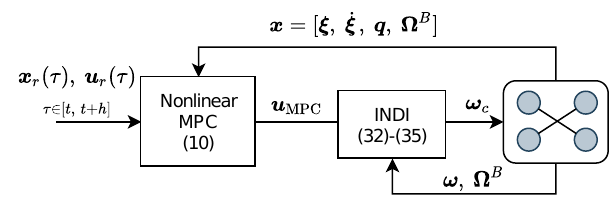}
    \caption{The control diagram of the model predictive controller with an INDI inner-loop controller.}
    \label{fig:diagram_MPC}
\end{figure*}

This section elaborates the two control methods compared. A nonlinear NMPC method is selected that considers the thrust limits of each rotor, the full nonlinear dynamics, and the aerodynamic effects. 
As for the DFBC method, we improve the state-of-the-art such that these factors are also addressed, which ensures a fair comparison with NMPC. 
Finally, both methods are augmented with an INDI controller~\cite{smeur2016adaptive} to convert the single rotor thrust commands to rotor-speed commands, while improving the robustness against model uncertainties and disturbances on the rotational dynamics. 
The control diagrams of both methods are illustrated in Fig.~\ref{fig:diagram_DFC} and Fig.~\ref{fig:diagram_MPC}.

\subsection{Nonlinear Model Predictive Controller}
NMPC generates control commands by solving a finite-time optimal control problem (OCP) in a receding horizon fashion. Given a reference trajectory, the cost function is the error between the predicted states and the reference states in the time horizon, meaning that multiple reference points in the time horizon are used. In order to conduct numerical optimizations, we discretize the states and inputs into $N$ equal intervals over the time horizon $\tau \in \left[t,~t+h\right]$ of size $dt=h/N$ \add{with $h$ denoting the horizon length}, yielding a constrained nonlinear optimization problem:
\begin{align}
\boldsymbol{u}_\mathrm{NMPC} =& \argmin_{\boldsymbol{u}}
\begin{multlined}
\sum_{k=0}^{N-1} \left(||\boldsymbol{x}_k - \boldsymbol{x}_{k,r}||^2_{\boldsymbol{Q}} + ||\boldsymbol{u}_k - \boldsymbol{u}_{k,r}||^2_{\boldsymbol{Q}_u} \right)
\\
 + ||\boldsymbol{x}_N - \boldsymbol{x}_{N,r}||_{\boldsymbol{Q_N}}
\end{multlined}\notag \\ 
& s.t. \quad \boldsymbol{x}_{k+1} = \boldsymbol{f}(\boldsymbol{x}_k, \boldsymbol{u}_k), \qquad \boldsymbol{x}_0 = \boldsymbol{x}_\mathrm{init}, \notag \\
& \qquad \boldsymbol{\Omega}^B \in \left[\boldsymbol{\Omega}^B_\mathrm{min}~\boldsymbol{\Omega}^B_\mathrm{max} \right], \quad \boldsymbol{u} \in \left[\boldsymbol{u}_\mathrm{min}~ \boldsymbol{u}_\mathrm{max}\right],
\label{eq:thrust_MPC}
\end{align}
Note that, we use the thrust of rotors $\boldsymbol{u}$ defined in (\ref{eq:equation_T_tau}) and (\ref{eq:equation_u}) as the control input of NMPC. The state vector is defined as $\boldsymbol{x} = \left[\boldsymbol{\xi}~\dot{\boldsymbol{\xi}}~\boldsymbol{q}~\boldsymbol{\Omega}^B\right]$, and
\begin{equation}
\boldsymbol{Q} = \mathrm{diag}\left(\boldsymbol{Q}_\xi,~\boldsymbol{Q}_v,~\boldsymbol{Q}_q,~\boldsymbol{Q}_\Omega\right),~\boldsymbol{Q}_N = \boldsymbol{Q}
\end{equation}
The reference state vector $\boldsymbol{x}_r$ and input $\boldsymbol{u}_r$ can be obtained from a trajectory planner \add{which generates full states} such as the one introduced in~\cite{foehn2021time}.
Function $\boldsymbol{f}\left(\boldsymbol{x}_k,\boldsymbol{u}_k\right)$ is the discretized version of nonlinear quadrotor model (\ref{eq:translate})-(\ref{eq:equation_u}). The same as many other works (see, e.g., \cite{bicego2020nonlinear, jacquet2020motor}), we omit $\boldsymbol{G}_2$ and $\boldsymbol{G}_3$ related terms in (\ref{eq:equation_T_tau}) as their effects are negligible. $\boldsymbol{x}_\mathrm{init}$ is the current state estimation when solving the OCP. $\boldsymbol{u}_\mathrm{min}$ and $\boldsymbol{u}_\mathrm{max}$ are minimum and maximum values of motor thrusts. Constraints on angular velocities are also added, which is found beneficial for improving the stability of NMPC according to~\cite{torrente2021data}. Note that in the above optimization problem, the following abuse of notation is used when calculating quaternion error: 
\begin{equation}
\boldsymbol{q} - \boldsymbol{q}_{r}  = \left( \boldsymbol{q}\otimes\boldsymbol{q}_{r}^{-1}\right)_{x,y,z}
\end{equation}

The above NMPC solves the full nonlinear model of a quadrotor, instead of resorting to a cascaded structure, or linear assumptions. 
In this paper, this quadratic nonlinear optimization problem is solved by a sequential quadratic programming (SQP) algorithm executed in a real-time iteration scheme~\cite{diehl2006fast}. We implement this algorithm using ACADO~\cite{verschueren2018towards} toolkit with qpOASES~\cite{ferreau2014qpoases} as the solver. More implementation details are given in Sec.~\ref{sec:experimental_setup}.
\subsection{Differential-Flatness Based Controller}

\begin{figure*}
    \centering
    \includegraphics[width=0.9\textwidth]{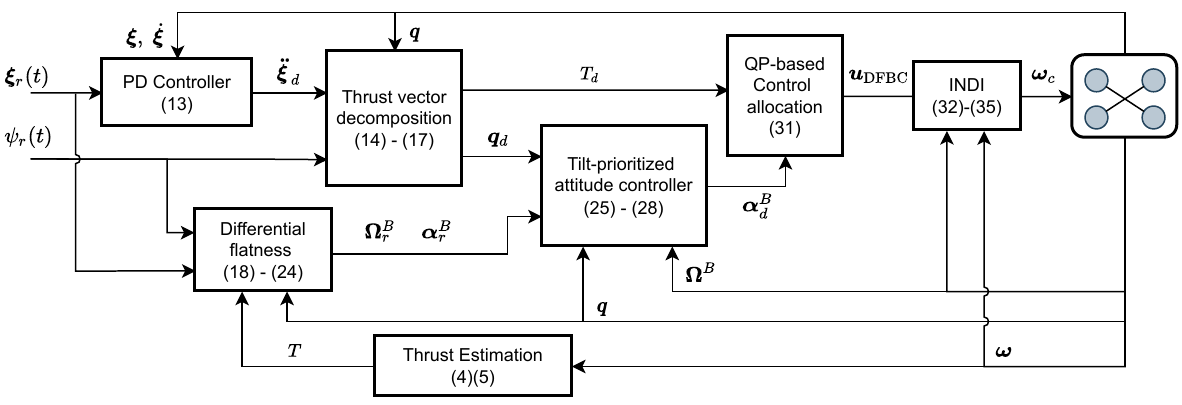}
    \caption{The control diagram of the differential-flatness-based controller, combined with an INDI inner-loop controller.}
    \label{fig:diagram_DFC}
\end{figure*}
Quadrotors are differentially flat systems~\cite{mellinger2011minimum}, namely, all the states and inputs can be written as algebraic functions of the flat outputs and their derivatives. 
\add{This allows a direct mapping from the flat outputs (positions $\boldsymbol{\xi}$ and heading $\psi$) to the angular rates and angular accelerations, which are leveraged by DFBC as feed-forward terms to improve the tracking accuracy.}

In the following, we introduced the DFBC method improved from a previous work~\cite{mellinger2011minimum}, where the original geometric attitude controller is replaced by the tilt-prioritized method~\cite{brescianini2018tilt}. We also use the quadratic-programming based control allocation~\cite{johansen2013control} to address input constraints. These modifications are  beneficial in tracking dynamically \textit{infeasible} trajectories, as will be discussed in Sec.~\ref{sec:tracking_infeasible_sim}. Fig.~\ref{fig:diagram_DFC} presents an overview of this method.

\subsubsection{Desired Attitude and Collective Thrust}
\add{First of all, we calculate the desired acceleration from a PD controller:
\begin{equation}
    \boldsymbol{\ddot{\xi}}_d = \boldsymbol{K}_\xi \left(\boldsymbol{\xi}_{r} - \boldsymbol{\xi}\right) + \boldsymbol{K}_v \left(\dot{\boldsymbol{\xi}}_{r} - \dot{\boldsymbol{\xi}}\right) + \ddot{\boldsymbol{\xi}}_r
    \label{eq:linear_control1}
\end{equation}
where $\boldsymbol{K}_\xi$ and $\boldsymbol{K}_v$ are positive-definite diagonal gain matrices. Then from (\ref{eq:translate}) and (\ref{eq:aero_force}), the desired thrust $T_d$ and thrust direction $\boldsymbol{z}_{B,d}$ are obtained as
\begin{equation}
    T_d\boldsymbol{z}_{B,d} = \left(\boldsymbol{\ddot{\xi}}_d - \boldsymbol{g}\right)m - \boldsymbol{R}(\boldsymbol{q})\boldsymbol{f}_a^B
    \label{eq:linear_control2}
\end{equation}}

Given reference heading angle $\psi_r$, we get an intermediate axis $\boldsymbol{x}_{C,d}$, by which the desired attitude can be obtained by the following equations:
\begin{equation}
\boldsymbol{x}_{C,d} = [\cos{\psi_{r}},~\sin{\psi_{r}},~0]^T,
\end{equation}
\begin{equation}
\boldsymbol{y}_{B,d} = \frac{\boldsymbol{z}_{B,d}\times\boldsymbol{x}_{C,d}}{||\boldsymbol{z}_{B,d}\times\boldsymbol{x}_{C,d}||},~\boldsymbol{x}_{B,d} = \boldsymbol{y}_{B,d} \times \boldsymbol{z}_{B,d}
\end{equation}
\begin{equation}
\boldsymbol{R}(\boldsymbol{q}_d) = [\boldsymbol{x}_{B,d},~\boldsymbol{y}_{B,d},~\boldsymbol{z}_{B,d}].
\end{equation}
where $\boldsymbol{q}_d$ is the desired attitude expressed by the quaternion.
\subsubsection{Reference Angular Velocity and Acceleration}
We leverage the differential flatness property of a quadrotor to derive the reference angular velocity and angular accelerations. Using them into the attitude control can help in tracking jerk and snap (3rd and 4th order derivatives of position $\boldsymbol{\xi}$), which is found beneficial to the tracking performance~\cite{tal2020accurate}.

Taking the derivative of both sides of (\ref{eq:translate}) and assuming a constant external aerodynamic force $\boldsymbol{f}_a$, we have
\begin{equation}
    m\dddot{\boldsymbol{\xi}} = \dot{T}\boldsymbol{z}_B + T\boldsymbol{\Omega}\times\boldsymbol{z}_B.
    \label{eq:kinematic_jerk}
\end{equation}
Then given reference jerk $\dddot{\boldsymbol{\xi}}_r$ and substitute it for $\dddot{\boldsymbol{\xi}}$ in (\ref{eq:kinematic_jerk}), we get
\add{
\begin{equation}
\boldsymbol{h}_\Omega \triangleq \boldsymbol{\Omega} \times \boldsymbol{z}_B = (m\dddot{\boldsymbol{\xi}}_{r} - \dot{T}\boldsymbol{z}_B) / T,
\label{eq:differential_flat_omega_h}
\end{equation}}
by which the reference angular velocity can be obtained as
\begin{equation}
\boldsymbol{\Omega}_{r}^B = \left[ -\boldsymbol{h}_\Omega \cdot \boldsymbol{y}_B,\qquad \boldsymbol{h}_\Omega \cdot \boldsymbol{x}_B, \qquad \dot{\psi}_{r}\boldsymbol{z}_I\cdot\boldsymbol{z}_B \right]^T,
\label{eq:differential_flat_omega}
\end{equation}
where $\psi_r$ is the reference heading angle.

We further take the derivative on both sides of (\ref{eq:kinematic_jerk}) and uses snap reference $\ddddot{\boldsymbol{\xi}}_r$ to replace $\ddddot{\boldsymbol{\xi}}$, yielding
\add{
\begin{multline}
    \boldsymbol{h}_{\boldsymbol{\alpha}} \triangleq \dot{\boldsymbol{\Omega}} \times \boldsymbol{z}_B \\
    = \frac{m}{T}\ddddot{\boldsymbol{\xi}}_r - \left(\boldsymbol{\Omega}\times(\boldsymbol{\Omega}\times\boldsymbol{z}_B) + \frac{2\dot{T}}{T}\boldsymbol{\Omega}\times\boldsymbol{z}_B + \frac{\ddot{T}}{T}\boldsymbol{z}_B\right).
    \label{eq:differential_flat_alpha_h}
\end{multline}}
Then the desired angular acceleration can be obtained as
\begin{equation}
    \boldsymbol{\alpha}_{r}^B = \left[ -\boldsymbol{h}_\alpha \cdot \boldsymbol{y}_B,\qquad \boldsymbol{h}_\alpha \cdot \boldsymbol{x}_B, \qquad \ddot{\psi}_{r}\boldsymbol{z}_I\cdot\boldsymbol{z}_B \right]^T.
    \label{eq:differential_flat_alpha}
\end{equation}
Note that in (\ref{eq:differential_flat_omega}) to (\ref{eq:differential_flat_alpha}) we use the current attitude $\left\{\boldsymbol{x}_B,~\boldsymbol{y}_B,~\boldsymbol{z}_B\right\}$, angular velocity $\boldsymbol{\Omega}$, and collective thrust $T$ instead of their references. Hence, the drone still follows the reference jerk $\dddot{\boldsymbol{\xi}}_r$ and snap $\ddddot{\boldsymbol{\xi}}_r$ even though its attitude, angular rates, and collective thrust have been deviated from the reference.

\add{Above derivations for $\boldsymbol{\Omega}^B_r$ and $\boldsymbol{\alpha}^B_r$ need the value of collective thrust $T$ and its derivatives}. While $T$ can be calculated from (\ref{eq:equation_T_tau})(\ref{eq:equation_u}) with measured rotor speed $\boldsymbol{\omega}$, its derivatives ($\dot{T}$ and $\ddot{T}$) are unable to be directly measured. For this reason, we approximate them by using reference jerk $\dddot{\boldsymbol{\xi}}_r$ and snap $\ddddot{\boldsymbol{\xi}}_r$. Multiplying (dot-product) both sides of (\ref{eq:kinematic_jerk}) by $\boldsymbol{z}_B$, we have
\begin{equation}
    \dot{T} = m\dddot{\boldsymbol{\xi}}_r \cdot \boldsymbol{z}_B
    \label{eq:T_dot}
\end{equation}
and its derivative
\begin{equation}
    \ddot{T} = m\ddddot{\boldsymbol{\xi}}_r \cdot \boldsymbol{z}_B + m(\boldsymbol{\Omega}\times\boldsymbol{z}_B)\cdot \dddot{\boldsymbol{\xi}}_r.
    \label{eq:T_ddot}
\end{equation}
\add{Equation (\ref{eq:T_dot}) and (\ref{eq:T_ddot}) are then substituted into (\ref{eq:kinematic_jerk})-(\ref{eq:differential_flat_alpha}) to calculate $\boldsymbol{\Omega}_r^B$ and $\boldsymbol{\alpha}_r^B$.}
\subsubsection{Tilt-Prioritized Attitude Control}
Quadrotors use rotor drag torques to achieve heading control. The control effectiveness of heading is around one order of magnitude lower than pitch and roll. As a consequence, heading control is prone to cause motor saturations. Fortunately, the thrust orientation of a quadrotor (namely tilt) is independent of its heading angle. Thus tilt-prioritized control has been proposed in~\cite{brescianini2018tilt} that regulates the reduced-attitude (pitch and roll) error $\Tilde{\boldsymbol{q}}_{e,red}$ and yaw error $\Tilde{\boldsymbol{q}}_{e,yaw}$ separately as follows:
\begin{equation}
[q_{e,w},~q_{e,x},~q_{e,y},~q_{e,z}]^T = \boldsymbol{q}_d \otimes \boldsymbol{q}^{-1},
\end{equation}

\begin{equation}
\Tilde{\boldsymbol{q}}_{e,red} = \frac{1}{\sqrt{q_{e,w}^2 + q_{e,z}^2}}\left[ \begin{array}{c}
      q_{e,w}q_{e,x} - q_{e,y}q_{e,z}\\
      q_{e,w}q_{e,y} + q_{e,x}q_{e,z} \\
      0
\end{array}\right],
\end{equation}
\begin{equation}
\Tilde{\boldsymbol{q}}_{e,yaw} = \frac{1}{\sqrt{q_{e,w}^2 + q_{e,z}^2}}\left[
 0 \quad 0 \quad q_{e,z}\right]^T.
\end{equation}
Subsequently, the desired angular acceleration can be obtained by the following attitude control law:
\begin{multline}
\boldsymbol{\alpha}_{d}^B = k_{q,red} \Tilde{\boldsymbol{q}}_{e,red} +  k_{q,yaw} \mathrm{sgn}(q_{e,w})\Tilde{\boldsymbol{q}}_{e,yaw}\\
+K_\Omega (\add{\boldsymbol{\Omega}_r^B - \boldsymbol{\Omega}^B}) + \boldsymbol{\alpha}^B_{r}
\label{eq:alpha_desired}
\end{multline}
where $k_{q,red}$ and $k_{e,yaw}$ are positive gains for reduced-attitude and yaw control respectively. Setting a relatively high $k_{q,red}$ over $k_{e,yaw}$ is helpful in improving position tracking accuracy while preventing input saturations. \add{$\boldsymbol{\alpha}^B_r$ in (\ref{eq:alpha_desired}) performs as a feed-forward term. It is worth noting that, the inclusion of $\boldsymbol{\alpha}^B_r$} seems to be reasonable from a theoretical perspective~\cite{brescianini2018tilt}, although removing it has been found to have almost no effect in our real-world experiments.

\subsubsection{Quadratic-Programming-Based Control Allocation}
The control allocation module generates thrust commands of each individual rotor from desired collective thrust $T_d$ and angular acceleration $\boldsymbol{\alpha}_d^B$. The same as NMPC, we also neglect the $\boldsymbol{G}_2$ and $\boldsymbol{G}_3$ terms in (\ref{eq:equation_T_tau}). Then from (\ref{eq:rot_dyn}) and (\ref{eq:equation_T_tau}), we obtain the direct-inversion control allocation:
\begin{align}
\boldsymbol{u} &= \boldsymbol{G}_1^{-1} \left[\begin{array}{c}
     T_d  \\
     \boldsymbol{I}_v \boldsymbol{\alpha}_{d}^B + \boldsymbol{\Omega}^B\times\boldsymbol{I}_v\boldsymbol{\Omega}^B
\end{array}\right], \\
    \boldsymbol{u}_\mathrm{DFBC} &= \mathrm{max}\left(\boldsymbol{u}_\mathrm{min},~\mathrm{min}(\boldsymbol{u},~ \boldsymbol{u}_\mathrm{max}) \right)
\label{eq:inversion_allocation}
\end{align}
This, however, does not consider input limits and may cause loss of control. For instance, an excessive collective thrust command can saturate all motors and consequently disable the attitude control. 

An effective alternative to address saturations is the quadratic-programming based allocation that solves the following optimization problem:
\begin{equation}
\begin{aligned}
\boldsymbol{u}_\mathrm{DFBC} =& \argmin_{\boldsymbol{u}} \quad \left\lVert\left[\begin{array}{c}
     T_d  \\
     \boldsymbol{I}_v \boldsymbol{\alpha}_{d}^B + \boldsymbol{\Omega}^B\times\boldsymbol{I}_v\boldsymbol{\Omega}^B
\end{array}\right] - \boldsymbol{G}_1\boldsymbol{u} \right\rVert_{\boldsymbol{W}} \\
& s.t. \quad \boldsymbol{u} \in \left[\boldsymbol{u}_\mathrm{min}~ \boldsymbol{u}_\mathrm{max}\right],
\end{aligned}
\label{eq:thrust_QP}
\end{equation} 
where $\boldsymbol{W}\in \mathbb{R}^{4\times 4}$ is a positive-definite diagonal weight matrix. Each diagonal entry respectively indicates the weight on the thrust, pitch, roll and yaw channels. Generally, setting a relatively high weight on pitch and roll (see Table.~\ref{tab:control_params}) is advantageous to prevent quadrotor loss-of-control when motor saturations are inevitable (e.g., tracking dynamically infeasible trajectories). If the solution is originally within control bounds, the result is the same as the direct-inversion allocation from (\ref{eq:inversion_allocation}). As for the implementation details, we solve this quadratic programming problem using an Active-Set Method from the qpOASES solver~\cite{ferreau2014qpoases}.

\subsection{Incremental Nonlinear Dynamic Inversion}
\label{sec:methodologies_indi}
After obtaining thrust commands from (\ref{eq:thrust_MPC}) or (\ref{eq:thrust_QP}), one can use (\ref{eq:equation_u}) to directly obtain the rotor speed command. 
However, the above-mentioned controllers \add{neglects the unmodeled term $\boldsymbol{d}_\tau$ in the rotational dynamics (\ref{eq:rot_dyn})}, which are found detrimental to the overall control performance. 

\add{Modeling $\boldsymbol{d}_\tau$ is very challenging for real-world systems.} Therefore, we resort to incremental nonlinear dynamic inversion (INDI), a sensor-based controller that uses instantaneous sensor measurement, instead of an explicit model, to represent system dynamics, thus being robust against model uncertainties and external disturbances. The performance and robustness of INDI have been demonstrated in previous research (see, e.g., \cite{smeur2016adaptive,tal2020accurate,sun2020incremental}) with proven stability~\cite{wang2019stability}. 

We use INDI as the inner-loop controller of both NMPC and DFBC for fair comparisons. The hybridization of INDI and DFBC is similar to a related work~\cite{tal2020accurate}, except for the attitude controller and control allocation introduced in the previous section. 
Here, a method is proposed to effectively combine INDI with NMPC to improve its robustness against rotational model uncertainties. 

After knowing the single rotor thrust command $\boldsymbol{u}_\mathrm{DFBC}$ or $\boldsymbol{u}_\mathrm{NMPC}$ from (\ref{eq:thrust_QP}) and (\ref{eq:thrust_MPC}) respectively, we can retrieve the desired collective thrust, and desired angular acceleration using (\ref{eq:rot_dyn}) and (\ref{eq:equation_T_tau}), yielding
\begin{equation}
\left[\begin{array}{c}
\hat{T}_d\\
\boldsymbol{I}_v\hat{\boldsymbol{\alpha}}_d^B + \boldsymbol{\Omega}^B\times\boldsymbol{I}_v\boldsymbol{\Omega}^B
\end{array}\right] = \boldsymbol{G}_1\boldsymbol{u}_\mathrm{DFBC/NMPC}.
\end{equation}
Note that, for the DFBC method, $\hat{T}_d$ and $\hat{\boldsymbol{\alpha}}^B$ are different from those derived from (\ref{eq:linear_control2}) and (\ref{eq:alpha_desired}) if the optimal cost of (\ref{eq:thrust_QP}) is non-zero. 
Then from INDI, we can get the desired body torque (see (30) and (31) of \cite{tal2020accurate} for detailed derivations)
\begin{equation}
\boldsymbol{\tau}_d = \boldsymbol{\tau}_f + \boldsymbol{I}_v\left(\hat{\boldsymbol{\alpha}}_d^B - \dot{\boldsymbol{\Omega}}^B_f\right)
\label{eq:tau_d}
\end{equation}
\add{Hence, the effect of unmodeled $\boldsymbol{d}_\tau$ is captured by filtered angular acceleration measurement $\dot{\boldsymbol{\Omega}}^B_f$ and filtered body torque $\boldsymbol{\tau}_f$, where} 
\begin{equation}
\boldsymbol{\tau}_f = \bar{\boldsymbol{G}}_1 c_t\boldsymbol{\omega}_f^{\circ 2} + \Delta t^{-1}\bar{\boldsymbol{G}}_2c_t \left(\boldsymbol{\omega}_f - \boldsymbol{\omega}_{f,k-1}\right)
\label{eq:tau_f}
\end{equation}
\add{is calculated from rotor speed measurements}. $\boldsymbol{\Omega}^B_f$ and $\boldsymbol{\omega}_f$ are \add{low-pass} filtered body-rate and rotor speeds \add{with the same cut-off frequencies. Thus they have a similar amount of delay and can be synchronized. In this paper, we use a second-order Butterworth filter with a 12~Hz cut-off frequency.}
Note that we use subscript $k-1$ to indicate the last sampled variable, and $\Delta t$ is the sampling interval. $\bar{\boldsymbol{G}}_1$ and $\bar{\boldsymbol{G}}_2$ respectively indicate matrices formed by the last three rows of $\boldsymbol{G}_1$ and $\boldsymbol{G}_2$. 

Then from (\ref{eq:linear_control2}), the following equation is obtained to solve rotor speed command $\boldsymbol{\omega}_c$:
\begin{equation}
\left[\begin{array}{c}
     \hat{T}_d  \\
     \boldsymbol{\tau} _d
\end{array}\right] = \boldsymbol{G}_1c_t \boldsymbol{\omega}_c^{\circ 2} + \Delta t^{-1}{\boldsymbol{G}}_2 c_t\left(\boldsymbol{\omega}_c - \boldsymbol{\omega}_{c,k-1}\right),
\label{eq:solve_omega_d}
\end{equation}
with the only unknown $\boldsymbol{\omega}_c$ which can be solved numerically. Differently from \cite{tal2020accurate}, the motor time constant is not needed in (\ref{eq:solve_omega_d}).

The INDI inner-loop controller converts the original thrust inputs from the high-level controller to rotor speed commands. 
Because this conversion  is through algebraic equations, system delay typically seen in the cascaded control structures can be effectively circumvented. 
The advantage of INDI over classical PID inner-loop controller will be demonstrated in Sec.~\ref{sec:real_world_experiments}.
\section{Implementation Details}\label{sec:experimental_setup}
Before elaborating on the simulation and real-world experiments, we introduce the implementation details and experimental setups. Both NMPC and DFBC flight controllers are implemented in an open-sourced flight stack \textit{Agilicious}\footnote{\url{https://agilicious.dev/}} programmed in C++, with ACADO~\cite{verschueren2018towards} toolkit and qpOASES~\cite{ferreau2014qpoases} as external libraries. A wrapper in ROS environment is also written which allows data logging, interfacing, and network communication. The controller gains are listed in Table~\ref{tab:control_params} and the inertial and geometric parameters of the tested quadrotor are listed in Table~\ref{tab:quads}. These parameters are used for both simulation and real-world experiments.

In the simulation experiments, the flight software runs on a laptop at 300~Hz while the frequency of NMPC is limited to 100~Hz for consistency with the real-world experiments. The rotor speed command is sent to a simulator included in \textit{Agilicious} that uses a 4th-order Runge-Kutta integrator to propagate quadrotor dynamic equations (\ref{eq:translate})-(\ref{eq:equation_T_tau}) at 500~Hz. To simulate the motor dynamics, rotor speed commands generated by the controllers pass through a low-pass filter with a 30~ms time constant. The drag model (\ref{eq:aero_force}) is also included to study the effect of aerodynamics, of which the parameters are included in Table~\ref{tab:quads}. The quadrotor states from the simulator are directly fed into the controllers unless a dedicated state estimation error is simulated.

The real-world experiment is performed in an indoor arena equipped with a motion capture system (VICON), with a 30 $\times$ 30 $\times$ 8~$\mathrm{m}^3$ tracking volume, as is shown in Fig.~\ref{fig:eye_catcher}. The flight software runs at 300~Hz on an onboard NVIDIA Jetson TX2 embedded computer. It includes the control algorithms (DFBC/NMPC and INDI inner-loop), and an extended Kalman filter fusing VICON (400~Hz) and IMU measurements (500~Hz) to obtain the full-state estimates. While DFBC runs at 300~Hz, the frequency of NMPC is limited to 100~Hz due to its computational complexity. The INDI inner-loop also runs at 300~Hz no matter which controller is being used. The onboard computer sends rotor speed commands of individual rotors to a low-level RadixFC board. The latter runs a customized firmware based on the open-source NuttX real-time operating system at 500~Hz to perform closed-loop rotor speed control, and sends rotor speed and IMU measurements to the onboard computer. We refer the users to \textit{Agilicious} for more details about the hardware.

\begin{table}[!htp]\centering
\caption{Controller gains and parameters}\label{tab:control_params}
\scriptsize
\begin{tabular}{ll|lll}\toprule
\multicolumn{2}{c|}{NMPC} &\multicolumn{2}{c}{DFBC} & \\\midrule
$\boldsymbol{Q}_\xi$ & diag(200, 200, 500) &$\boldsymbol{K}_\xi$ &diag(10, 10, 10) \\
$\boldsymbol{Q}_v$ &diag(1, 1, 1) &$\boldsymbol{K}_v$ &diag(6, 6, 6) \\
$\boldsymbol{Q}_q$ &diag(5, 5, 200) &($k_{q,red}, k_{q,yaw})$ &(150, 3) \\
$\boldsymbol{Q}_\Omega$ &diag(1, 1, 1) &$\boldsymbol{K}_\Omega$ &diag(20, 20, 8) \\
$\boldsymbol{Q}_u$ &diag(6, 6, 6, 6) &$\boldsymbol{W}$ &diag(0.001, 10, 10, 0.1) \\
$dt$ & 50 ms & &\\
$N$ & 20 & &\\
\bottomrule
\end{tabular}
\end{table}

\begin{table}[htp]
    \centering
    \setlength{\tabcolsep}{3pt}
    \caption{Quadrotor Configurations}
    \label{tab:quads}
    {\footnotesize
    \begin{tabular}{cc|c}
        \toprule
        \multicolumn{2}{c|}{Parameter(s)} & Value(s)\\
        \midrule
        $m$& $[\si{\kilo\gram}]$ & $0.75$ \\
        $l$& $[\si{\meter}]$ & $0.14$ \\
        $\beta$& $[\si{\deg}]$ & $56$ \\
        $\boldsymbol{I}_v$& $[\si{\gram\meter^2}]$ & $\mathrm{diag}(2.5, 2.1, 4.3)$\\
        $\left(u_\mathrm{min}, u_\mathrm{max}\right)$& $[\si{\newton}]$ & $\left(0, 8.5\right)$\\
        $c_q$& $[\si{Nm^2}]$ & $2.37e^{-8}$\\
        $c_t$& $[\si{N^2}]$ & $1.51e^{-6}$\\
        $\left(k_{d,x},~k_{d,y},~k_{d,z}\right)$& $[\si{kg/s}]$ & $\left(0.26,~0.28,~0.42\right)$\\
        $k_{h}$& $[\si{kg/m}]$ & $0.01$\\
        \bottomrule
    \end{tabular}
    }
    \vspace{-8pt}
\end{table}
\section{Simulation Experiments}
\label{sec:simulation_experiments}
In a set of controlled experiments in simulation, we examine DFBC and NMPC to address the following research questions:
(i) Given the computational budget and a well-identified model, which control approach achieves better tracking performance? (ii) How do these two approaches perform in tracking dynamically \textit{infeasible} trajectories, which inevitably lead to actuator saturation? 
(iii) How do these two approaches perform in simulated practical situations with a model mismatch, external disturbances, state estimation latency? 
Note that in this section, both methods are tested with the augmentation of the INDI inner-loop controller. 
In this section, we use NMPC and DFBC to represent those with INDI and an aerodynamic drag model for readability. 


\subsection{Reference Trajectories}
Multiple elliptical reference trajectories on both horizontal and vertical planes are generated for testing the tracking performance. \add{The results of tracking these simple trajectory primitives can further reflect the performance of tracking more complex 3D trajectories. These trajectories are parameterized by the maximum velocity $V_\mathrm{max}$, maximum acceleration $a_\mathrm{max}$, and ellipticity $n$.} Specifically, the horizontal reference trajectories are expressed as:
\begin{equation}
    \boldsymbol{\xi}_r(t) = \left[r_\mathrm{max}
 \sin(k t),~r_\mathrm{min}
  \cos(k t),~ 5 \right]^T;
\end{equation}
and the vertical reference trajectories are
\begin{equation}
    \boldsymbol{\xi}_r(t) = \left[r_\mathrm{max} \sin(k t),~0,~5+r_\mathrm{min}\cos(k t) \right]^T
\end{equation}
where
\begin{equation}
r_\mathrm{max} = V_\mathrm{max}^2 / a_\mathrm{max},~k=a_\mathrm{max}/V_\mathrm{max},~r_\mathrm{min} = r_\mathrm{max} / n.
\end{equation}
The heading angle of each horizontal reference trajectory is selected to always point at the center of the ellipse; for vertical trajectories, the reference heading angles are kept constant. Specifically, 144 reference trajectories are generated by combining parameters $a_\mathrm{max} \in \{10,~20,~30,~40,~50,~60\}~\mathrm{m/s^2}$, $V_\mathrm{max} \in \{5,~10,~15,~20\}~\mathrm{m/s}$,~$n\in \{1,~2,~5\}$. 

\add{Tracking some of these trajectories may require the quadrotor to exceed its thrust limitations. These trajectories are referred to as dynamically \textit{infeasible}. In many time-critical applications, such as autonomous drone racing, rotors need to reach their full thrust limits to fully exploit the agility of the platform and achieve faster lap times. Designing such an agile time-optimal trajectory requires a perfect model that captures real thrust limits and aerodynamic effects in high-speed flights. 
Without this model, the generated trajectory may exceed the quadrotor capabilities and becomes dynamically \textit{infeasible} to accurately follow.
Since such an accurate model is usually unattainable, studying the performance when tracking these \textit{infeasible} trajectories is necessary.}

\add{For the simulated quadrotor with the configurations given in Table~\ref{tab:quads}, there are 68 \textit{infeasible} trajectories and 76 \textit{feasible} trajectories. They are determined by being tracked by a modified NMPC without thrust limits imposed. If the applied thrust exceeds the maximum thrust of the real drone, the trajectory is marked as \textit{infeasible}. The feasibility of each reference trajectory is given in Table~\ref{tab:feasibility}}

\begin{table}[!htp]\centering
\caption{\add{Dynamical feasibility of reference trajectories. (\textit{feasible}:\ding{51}; \textit{infeasible}:\ding{55})}}\label{tab:feasibility}
\scriptsize
\begin{subtable}[h]{0.5\textwidth}
\caption{Horizontal trajectories.}
\begin{tabular}{c|cccc|cccc|ccccc}\toprule
\multirow{3}{*}{\makecell{$a_\mathrm{max}$ \\ $[$m/s$^2]$}} &\multicolumn{4}{c}{$n=1$} &\multicolumn{4}{c}{$n=2$} &\multicolumn{4}{c}{$n=5$} \\\cmidrule{2-13}
&\multicolumn{4}{c}{$V_\mathrm{max}$ [m/s]} &\multicolumn{4}{c}{$V_\mathrm{max}$ [m/s]} &\multicolumn{4}{c}{$V_\mathrm{max}$ [m/s]} \\\cmidrule{2-13}
&5 &10 &15 &20 &5 &10 &15 &20 &5 &10 &15 &20 \\\midrule
10 &\ding{51} &\ding{51} &\ding{51} &\ding{51} &\ding{51} &\ding{51} &\ding{51} &\ding{51} &\ding{51} &\ding{51} &\ding{51} &\ding{51} \\
20 &\ding{51} &\ding{51} &\ding{51} &\ding{51} &\ding{51} &\ding{51} &\ding{51} &\ding{51} &\ding{55} &\ding{51} &\ding{51} &\ding{51} \\
30 &\ding{51} &\ding{51} &\ding{51} &\ding{51} &\ding{55} &\ding{51} &\ding{51} &\ding{51} &\ding{55} &\ding{55} &\ding{51} &\ding{51} \\
40 &\ding{51} &\ding{51} &\ding{51} &\ding{51} &\ding{55} &\ding{51} &\ding{51} &\ding{51} &\ding{55} &\ding{55} &\ding{55} &\ding{51} \\
50 &\ding{55} &\ding{55} &\ding{55} &\ding{55} &\ding{55} &\ding{55} &\ding{55} &\ding{55} &\ding{55} &\ding{55} &\ding{55} &\ding{55} \\
60 &\ding{55} &\ding{55} &\ding{55} &\ding{55} &\ding{55} &\ding{55} &\ding{55} &\ding{55} &\ding{55} &\ding{55} &\ding{55} &\ding{55} \\
\bottomrule
\end{tabular}
\end{subtable}

\begin{subtable}[h]{0.5\textwidth}
\caption{Vertical trajectories.}
\begin{tabular}{c|cccc|cccc|ccccc}\toprule
\multirow{3}{*}{\makecell{$a_\mathrm{max}$ \\ $[$m/s$^2]$}} &\multicolumn{4}{c}{$n=1$} &\multicolumn{4}{c}{$n=2$} &\multicolumn{4}{c}{$n=5$} \\\cmidrule{2-13}
&\multicolumn{4}{c}{$V_\mathrm{max}$ [m/s]} &\multicolumn{4}{c}{$V_\mathrm{max}$ [m/s]} &\multicolumn{4}{c}{$V_\mathrm{max}$ [m/s]} \\\cmidrule{2-13}
&5 &10 &15 &20 &5 &10 &15 &20 &5 &10 &15 &20 \\\midrule
10 &\ding{51} &\ding{51} &\ding{51} &\ding{51} &\ding{51} &\ding{51} &\ding{51} &\ding{51} &\ding{51} &\ding{51} &\ding{51} &\ding{51} \\
20 &\ding{51} &\ding{51} &\ding{51} &\ding{51} &\ding{51} &\ding{51} &\ding{51} &\ding{51} &\ding{51} &\ding{51} &\ding{51} &\ding{51} \\
30 &\ding{51} &\ding{51} &\ding{51} &\ding{51} &\ding{51} &\ding{51} &\ding{51} &\ding{51} &\ding{51} &\ding{51} &\ding{51} &\ding{51} \\
40 &\ding{55} &\ding{55} &\ding{55} &\ding{55} &\ding{55} &\ding{55} &\ding{55} &\ding{55} &\ding{55} &\ding{55} &\ding{55} &\ding{55} \\
50 &\ding{55} &\ding{55} &\ding{55} &\ding{55} &\ding{55} &\ding{55} &\ding{55} &\ding{55} &\ding{55} &\ding{55} &\ding{55} &\ding{55} \\
60 &\ding{55} &\ding{55} &\ding{55} &\ding{55} &\ding{55} &\ding{55} &\ding{55} &\ding{55} &\ding{55} &\ding{55} &\ding{55} &\ding{55} \\
\bottomrule
\end{tabular}
\end{subtable}
\end{table}

\subsection{Evaluation Criteria}
In the following comparisons, we use the root mean square error (RMSE) of position and heading as the precision metric of a method. 
The crash rate is another criterion to show the robustness of a method. 
A certain flight is defined as \textit{crashed} if its position $\boldsymbol{\xi}$ violates the following spatial constraint at an arbitrary instant:
\begin{equation}
\mathrm{inf}\{\boldsymbol{\xi}_{r}(t)\} - \boldsymbol{b} \leq \boldsymbol{\xi} \leq \mathrm{sup}\{\boldsymbol{\xi}_{r}(t)\} + \boldsymbol{b}. 
\end{equation}
We select $\boldsymbol{b} = [5,~5,~5]^T$ meters in this study.

\subsection{Tracking Dynamically \textit{Feasible} Trajectories}
First of all, we compare the performance of NMPC and DFBC in tracking 76 \textit{feasible} trajectories. We perform the test in an ideal condition with perfect model knowledge and state estimates. Since only \textit{feasible} trajectories are tracked, there is no saturation of single rotor thrusts and both methods succeed in tracking all trajectories without a single crash. 

\begin{figure}
\centering
\subfloat[]{
	\includegraphics[width=0.49\textwidth]{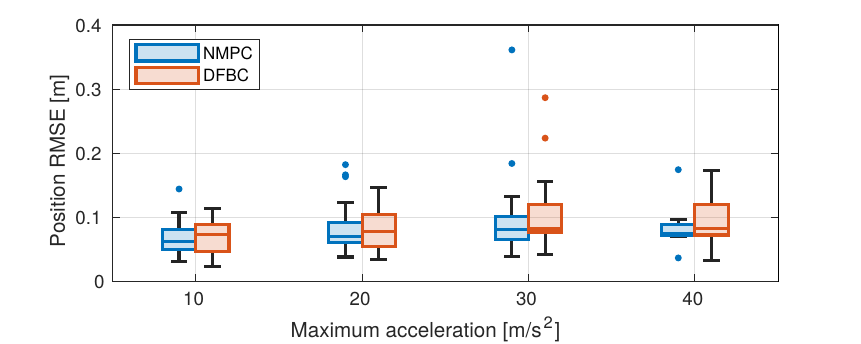} } \\
\subfloat[]{
	\includegraphics[width=0.49\textwidth]{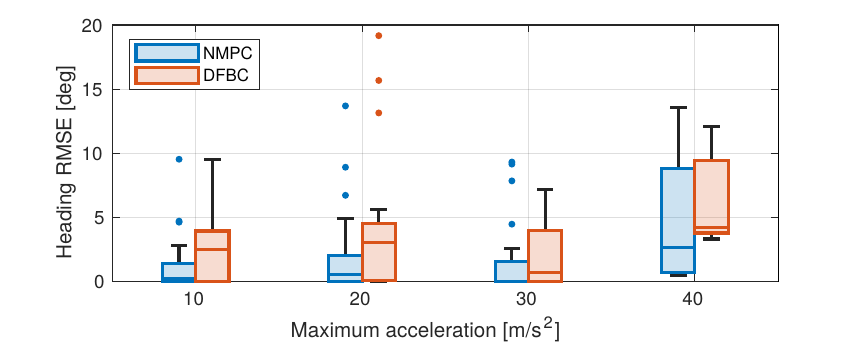} } 
	\caption{Boxplot of tracking error (RMSE) in tracking different dynamically \textit{feasible} trajectories \add{(76 in total)} categorized by maximum accelerations.}
	\label{fig:RMSE_no_sat}
\end{figure}

In this set of tests, we fine-tune the parameters of both methods (listed in Table.~\ref{tab:control_params}) \add{such that they achieve a similar position tracking error}. Fig.~\ref{fig:RMSE_no_sat} compares the boxplots of NMPC and DFBC when tracking trajectories with different reference maximum \add{accelerations}. Both methods have \add{similar} position RMSE in these flights. As for the heading tracking, NMPC has an average heading RMSE of \add{2.0}~\si{deg}, which is better than \add{5.8}~\si{deg} of DFBC. 

\begin{figure}
\centering
\subfloat[]{
	\includegraphics[width=0.49\textwidth]{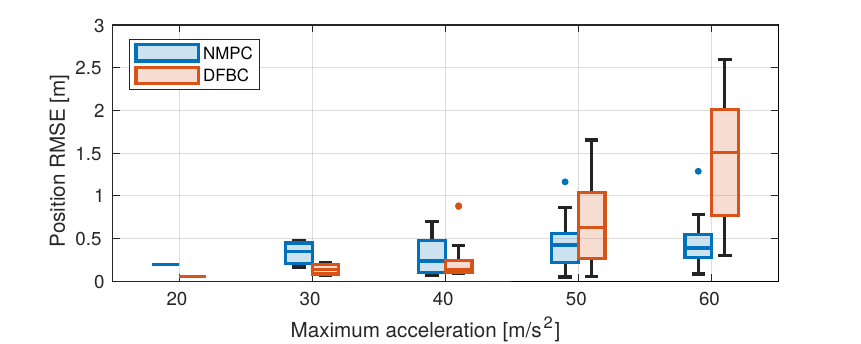} } \\
\subfloat[]{
	\includegraphics[width=0.49\textwidth]{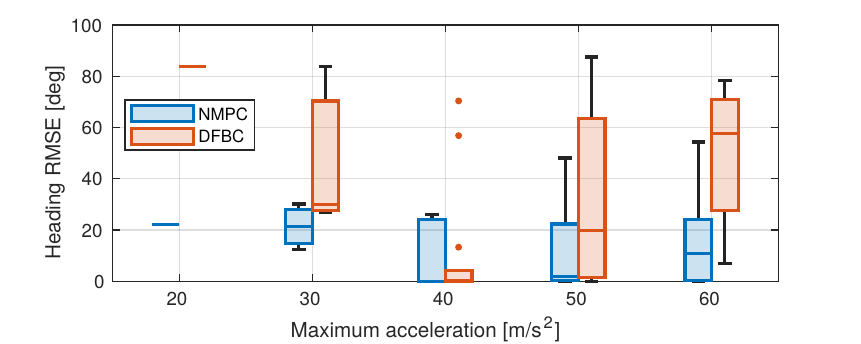} } 
	\caption{Boxplot of tracking error (RMSE) in tracking different dynamically \textit{infeasible} trajectories \add{(68 in total)} categorized by maximum accelerations.  Crashed flights are removed from these analyses (see Fig.~\ref{fig:crashRates_nonMPC} for the crash rate).}
	\label{fig:RMSE_with_sat}
\end{figure}

\subsection{Tracking Dynamically Infeasible Trajectories}
\label{sec:tracking_infeasible_sim}
\subsubsection{Tracking Accuracy}
Fig.~\ref{fig:RMSE_with_sat} shows the box plot of position and heading RMSE of NMPC and DFBC in tracking the previously generated 68 \textit{infeasible} trajectories. Crashed flights are excluded from this plot (see Fig.~\ref{fig:crashRates_nonMPC} for the crash rate). 
We find that the DFBC method shows smaller position RMSE in tracking trajectories with lower acceleration. 
However, as the reference acceleration grows, NMPC significantly outperforms DFBC. In general, NMPC outperforms DFBC by \add{48}\% on position tracking (\add{0.40}~\si{m} vs. \add{0.77}~\si{m}).
As for the heading tracking, NMPC has a noticeably less (\add{62}\%) RMSE than the DFBC method (\add{12.7}~\si{deg} vs. \add{33.4}~\si{deg}).
The advantage of NMPC over DFBC becomes particularly relevant if the reference acceleration exceeds the maximum thrust ($\sim$ 5g) of the tested quadrotor.

\subsubsection{Crash Rates}
\begin{figure}
    \centering
    \includegraphics[width=0.49\textwidth]{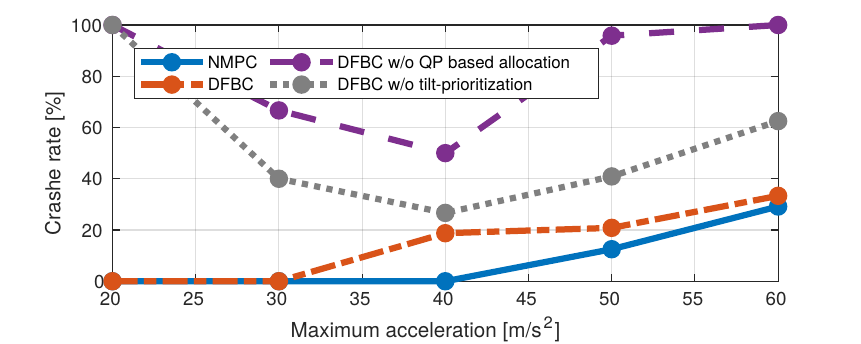}
    \caption{Crash rates of NMPC, DFBC, and DFBC with different setups: DFBC without quadratic-programming based allocation, DFBC without tilt-prioritized attitude controller but using the geometric attitude controller.}
    \label{fig:crashRates_nonMPC}
\end{figure}

Fig.~\ref{fig:crashRates_nonMPC} compares the crash rates of DFBC and NMPC in tracking all the \textit{infeasible} trajectories. NMPC (solid-blue) outperforms DFBC (red-dash-dot), especially in tracking trajectories with high reference accelerations. 

Here, we also demonstrate contributions of the tilt-prioritized attitude controller and the quadratic-programming based allocation presented in Sec.~\ref{sec:methodologies}. We replace them respectively with the conventional \textit{geometric attitude controller}~\cite{lee2010geometric} and the \textit{direct-inversion allocation} (\ref{eq:inversion_allocation}). The resultant crash rates are shown in Fig.~\ref{fig:crashRates_nonMPC} as well. It is obvious that the DFBC method tested in this study shows a significantly lower crash rate in tracking \textit{infeasible} trajectories.

\subsection{Robustness Study}

\begin{table*}[!htp]\centering
\caption{Tracking performance comparisons in different types of non-ideal conditions commonly seen in practice  \textbf{without INDI inner loop.} \add{Each number shows the mean and standard deviation (SD) across tracking 76 dynamically \textit{feasible} trajectories.}}\label{tab:robustness_study_wo_indi}
\scriptsize
\begin{tabular}{l|cc|cc|ccc}\toprule
&\multicolumn{2}{c|}{\makecell{ Position RMSE $[$m$]$ \\ \add{(mean $\pm$ SD)}}} &\multicolumn{2}{c|}{\makecell{Heading RMSE $[$deg$]$ \\ (mean $\pm$ SD)}} &\multicolumn{2}{c}{\makecell{Crash Rate \\ $[$\%$]$}} \\\cmidrule{2-7}
&NMPC &DFBC &NMPC &DFBC &NMPC &DFBC \\\midrule
 Baseline        & 0.084 $\pm$ 0.049 & 0.084 $\pm$ 0.041 &      \textbf{2.13} $\pm$      3.38 &      5.91 $\pm$     11.90  &     0   &     0   \\\midrule
 +50\% Drag   & \textbf{0.183} $\pm$ 0.066 & 0.194 $\pm$ 0.084 &      \textbf{2.11} $\pm$      3.35 &      8.56 $\pm$     15.46 &     0   &     0   \\
 +100\% Drag   & \textbf{0.286} $\pm$ 0.109 & 0.305 $\pm$ 0.139 &      \textbf{2.25} $\pm$      3.38 &      9.07 $\pm$     11.44 &     0   &     0   \\\midrule
 -30\% Mass      & \textbf{0.320}  $\pm$ 0.105 & 0.477 $\pm$ 0.109 &      \textbf{2.27} $\pm$      3.38 &     11.26 $\pm$     17.77 &     0   &     0   \\
 +30\% Mass       & 0.809 $\pm$ 1.455 & \textbf{0.687} $\pm$ 1.001 &     11.77 $\pm$     27.21 &      \textbf{7.21} $\pm$     14.11 &    10.7 &     \textbf{0}   \\\midrule
10\% CoG bias      & \textbf{0.264} $\pm$ 0.094 & 1.499 $\pm$ 1.802 &      \textbf{2.67} $\pm$      3.42 &     25.95 $\pm$     35.69 &     \textbf{0}   &    20   \\
 15\% CoG bias     & \textbf{0.464} $\pm$ 0.557 & 3.039 $\pm$ 2.216 &      \textbf{4.58} $\pm$     10.94 &     53.62 $\pm$     41.95 &     \textbf{1.3} &    56   \\\midrule
 -30\% $c_t$ & \textbf{0.939} $\pm$ 1.415 & 0.998 $\pm$ 1.216 &     12.30  $\pm$     27.13 &      \textbf{8.79} $\pm$     18.01 &    10.7 &     \textbf{0}   \\
 +30\% $c_t$  & \textbf{0.245} $\pm$ 0.085 & 0.36  $\pm$ 0.076 &      \textbf{2.05} $\pm$      3.34 &      6.63 $\pm$     10.87 &     0   &     0   \\\midrule
 -30\% $c_q$& 0.085 $\pm$ 0.052 & \textbf{0.084} $\pm$ 0.041 &      \textbf{2.03} $\pm$      3.37 &      6.23 $\pm$     13.03 &     0   &     0   \\
 +30\% $c_q$& 0.084 $\pm$ 0.051 & 0.084 $\pm$ 0.04  &      \textbf{2.07} $\pm$      3.38 &      5.81 $\pm$     11.81 &     0   &     0   \\\midrule
 5N Ext. force        & \textbf{0.252} $\pm$ 0.042 & 0.302 $\pm$ 0.031 &      \textbf{5.08} $\pm$      3.58 &     17.42 $\pm$     14.59 &     0   &     0   \\
10N Ext. force       & 0.716 $\pm$ 1.019 & \textbf{0.689} $\pm$ 0.524 &     \textbf{12.59} $\pm$     19.07 &     24.59 $\pm$     16.28 &     5.3 &     \textbf{1.3} \\
 15N Ext. force      & 1.870  $\pm$ 1.892 & \textbf{1.391} $\pm$ 1.069 &     \textbf{32.13} $\pm$     35.38 &     32.54 $\pm$     20.58 &    26.7 &     \textbf{6.7} \\\midrule
  0.1Nm Ext. moment        & \textbf{0.153} $\pm$ 0.038 & 0.270  $\pm$ 0.098 &      \textbf{3.26} $\pm$      3.35 &     12.04 $\pm$     13.45 &     0   &     0   \\
 0.2Nm Ext. moment       & \textbf{0.319} $\pm$ 0.545 & 2.538 $\pm$ 2.079 &      \textbf{5.85} $\pm$     10.71 &     54.89 $\pm$     31.08 &     \textbf{1.3} &    41.3 \\
 0.3Nm Ext. moment        & \textbf{0.566} $\pm$ 0.907 & 4.949 $\pm$ 0.439 &     \textbf{10.87} $\pm$     17.27 &     89.16 $\pm$      6.88 &     \textbf{4.0}   &    98.7 \\\midrule
 10ms Latency    & \textbf{0.085} $\pm$ 0.065 & 0.097 $\pm$ 0.099 &      \textbf{2.07} $\pm$      3.34 &      5.79 $\pm$     11.05 &     0   &     0   \\
 30ms Latency    & \textbf{0.158} $\pm$ 0.183 & \textbf{0.333} $\pm$ 0.976 &      \textbf{5.04} $\pm$     10.47 &     13.73 $\pm$     25.87 &     \textbf{0}   &     4.0   \\
 50ms Latency    & 3.450  $\pm$ 2.217 & \textbf{0.669} $\pm$ 1.247 &     65.00    $\pm$     38.65 &     \textbf{54.82} $\pm$     30.02 &    66.7 &     \textbf{5.3} \\
\bottomrule
\end{tabular}
\end{table*}

\begin{table*}[!htp]\centering
\caption{Tracking performance comparisons in different types of non-ideal conditions commonly seen in practice \textbf{with INDI inner loop.} \add{Each number shows the mean and standard deviation (SD) across tracking 76 dynamically \textit{feasible} trajectories.}}\label{tab:robustness_study}
\scriptsize
\begin{tabular}{l|cc|cc|ccc}\toprule
&\multicolumn{2}{c|}{\makecell{ Position RMSE $[$m$]$ \\ \add{(mean $\pm$ SD)}}} &\multicolumn{2}{c|}{\makecell{Heading RMSE $[$deg$]$ \\ (mean $\pm$ SD)}} &\multicolumn{2}{c}{\makecell{Crash Rate \\ $[$\%$]$}} \\\cmidrule{2-7}
&NMPC &DFBC &NMPC &DFBC &NMPC &DFBC \\\midrule
Baseline            & \textbf{0.082} $\pm$ 0.047 & 0.085 $\pm$ 0.043 &      \textbf{2.03} $\pm$      3.35 &      5.83 $\pm$     11.40  &     0   &     0   \\\midrule
+50\% Drag          & \textbf{0.184} $\pm$ 0.067 & 0.194 $\pm$ 0.084 &      \textbf{2.10}  $\pm$      3.33 &      8.48 $\pm$     14.50  &     0   &     0   \\
+100\% Drag         & \textbf{0.288} $\pm$ 0.11  & 0.306 $\pm$ 0.139 &      \textbf{2.25} $\pm$      3.32 &      9.32 $\pm$     11.99 &     0   &     0   \\\midrule
-30\% Mass          & \textbf{0.320}  $\pm$ 0.106 & 0.477 $\pm$ 0.108 &      \textbf{2.26} $\pm$      3.36 &     10.42 $\pm$     14.19 &     0   &     0   \\
+30\% Mass          & 0.747 $\pm$ 1.372 & \textbf{0.686} $\pm$ 0.999 &     10.60  $\pm$     25.66 &      \textbf{7.11} $\pm$     13.97 &     9.3 &     \textbf{0}   \\\midrule
10\% CoG bias       & \textbf{0.081} $\pm$ 0.047 & 0.099 $\pm$ 0.104 &      \textbf{2.09} $\pm$      3.32 &      6.46 $\pm$     13.65 &     0   &     0   \\
15\% CoG bias       & \textbf{0.083} $\pm$ 0.051 & 0.378 $\pm$ 1.117 &      \textbf{2.05} $\pm$      3.30  &     10.70  $\pm$     22.52 &     \textbf{0}   &     5.3 \\\midrule
-30\% $c_t$         & 1.362 $\pm$ 1.764 & \textbf{1.004} $\pm$ 1.212 &     19.98 $\pm$     33.88 &     \textbf{10.06} $\pm$     19.06 &    18.7 &     \textbf{1.3} \\
+30\% $c_t$         & \textbf{0.244} $\pm$ 0.085 & 0.359 $\pm$ 0.076 &      \textbf{2.13} $\pm$      3.52 &      6.14 $\pm$      9.58 &     0   &     0   \\\midrule
-30\% $c_q$      & \textbf{0.082} $\pm$ 0.049 & 0.086 $\pm$ 0.047 &      \textbf{2.08} $\pm$      3.35 &      6.01 $\pm$     11.69 &     0   &     0   \\
+30\% $c_q$      & \textbf{0.082} $\pm$ 0.046 & 0.085 $\pm$ 0.046 &      \textbf{2.07} $\pm$      3.34 &      5.83 $\pm$     11.61 &     0   &     0   \\\midrule
5N Ext. force       & \textbf{0.254} $\pm$ 0.051 & 0.302 $\pm$ 0.033 &      \textbf{5.03} $\pm$      3.48 &     17.58 $\pm$     14.77 &     0   &     0   \\
10N Ext. force      & 0.725 $\pm$ 1.018 & \textbf{0.690}  $\pm$ 0.524 &     \textbf{12.38} $\pm$     19.02 &     24.81 $\pm$     16.28 &     5.3 &     \textbf{1.3} \\
15N Ext. force      & 1.872 $\pm$ 1.891 & \textbf{1.393} $\pm$ 1.070  &     \textbf{32.14} $\pm$     35.35 &     32.97 $\pm$     20.80  &    26.7 &     \textbf{6.7} \\\midrule
0.1Nm Ext. moment   & \textbf{0.083} $\pm$ 0.048 & 0.087 $\pm$ 0.049 &      \textbf{2.49} $\pm$      3.22 &      7.95 $\pm$     12.46 &     0   &     0   \\
0.2Nm Ext. moment   & \textbf{0.087} $\pm$ 0.051 & 0.113 $\pm$ 0.088 &      \textbf{3.19} $\pm$      3.35 &     13.65 $\pm$     17.25 &     0   &     0   \\
0.3Nm Ext. moment   & \textbf{0.298} $\pm$ 0.962 & 0.764 $\pm$ 1.663 &      \textbf{8.40}  $\pm$     17.29 &     26.93 $\pm$     29.41 &     \textbf{4.0}   &    13.3 \\\midrule
10ms Latency        & \textbf{0.087} $\pm$ 0.088 & 0.099 $\pm$ 0.106 &      \textbf{2.21} $\pm$      3.44 &      5.73 $\pm$     11.65 &     0   &     0   \\
30ms Latency        & 0.280  $\pm$ 0.802 & \textbf{0.265} $\pm$ 0.800   &      \textbf{8.39} $\pm$     20.12 &     14.34 $\pm$     25.85 &     2.7 &     2.7 \\
50ms Latency        & 3.480  $\pm$ 2.235 & \textbf{0.694} $\pm$ 1.294 &     64.02 $\pm$     39.96 &     \textbf{52.29} $\pm$     29.12 &    68.0   &     \textbf{6.7} \\
\bottomrule
\end{tabular}
\end{table*}
While previous simulations are carried out in an ideal condition, this section studies the robustness of NMPC and DFBC methods in the following non-ideal conditions:
\begin{itemize}
    \item Model mismatch, including the center of gravity bias, mass mismatch, thrust and torque coefficients model mismatch, and error of aerodynamic drag model.
    \item External disturbances. In details, we simulate constant external forces along $\boldsymbol{x}_I$, or external torques along both $\boldsymbol{x}_B$ and $\boldsymbol{y}_B$. These disturbances last for 5 seconds.
    \item Position, velocity, and attitude estimation latency. This aims at studying the effect of latency from such as signal transmission, state estimation algorithms, and sensors. As angular rates are measured directly by IMU with negligible latency, we only study the latency on pose and velocity estimates.
\end{itemize}
These robustness studies are conducted using only the \textit{feasible} trajectories since NMPC already shows its advantage over DFBC in tracking \textit{infeasible} trajectories.
We set the position RMSE as 5.0~m and heading RMSE as 90~deg for those crashed flights. 

Table~\ref{tab:robustness_study_wo_indi} and Table~\ref{tab:robustness_study} respectively present the results without and with INDI inner-loop. NMPC shows substantially higher robustness than DFBC against model uncertainties and disturbances on rotational dynamics due to CoG bias or external moments. Adding an INDI inner-loop controller can improve the robustness against these uncertainties.
Even so, NMPC still slightly outperforms DFBC when both are augmented by an INDI inner-loop controller.

\begin{figure}
    \centering
    \includegraphics[width=0.49\textwidth]{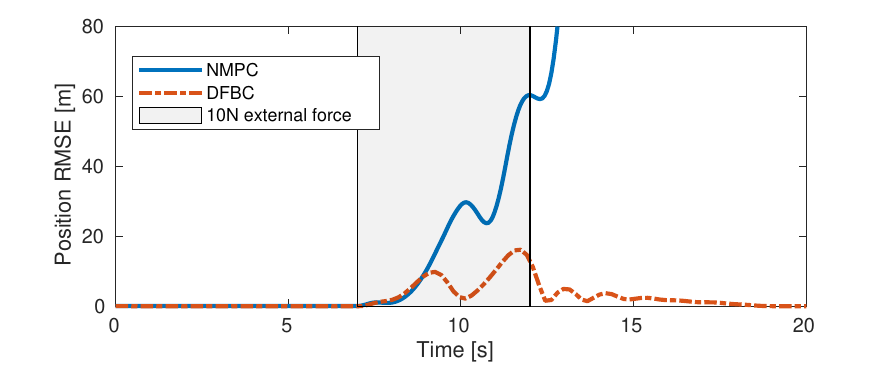}
    \caption{Position tracking error for a loop trajectory ($V_{max} = 15m/s$, $a_{max}=40\mathrm{m/s^2}$, $n=1$). Grey area indicates the period with 10~N lateral disturbances acting on the drone. NMPC failed to converge and crashed the drone, while the DFBC method succeeds in recover\add{ing} the drone after the external disturbance disappeared.}
    \label{fig:MPC_crash_disturbance}
\end{figure}

Unlike the rotational dynamics, NMPC shows a higher crash rate while experiencing model uncertainties and disturbances acting on the translational dynamics. 
For example, when the real mass is 30\% higher, both controllers generate fewer thrusts than required to track the trajectory, resulting in a constant position tracking error.
With this tracking error, NMPC fails to converge in over 10\% of flights and crashes the drone. 
The same reason also explains the significantly higher crash rate of NMPC in the presence of large external disturbances. Fig.~\ref{fig:MPC_crash_disturbance} shows an example where NMPC failed in converging and crashed the drone after experiencing a 10~N external force disturbance. \add{Note that adding INDI inner-loop does not improve the performance of either controller to reject disturbances in the translational dynamics. A previous research~\cite{tal2020accurate} implemented INDI in the position control (INDI outer-loop) for DFBC by virtue of its cascaded structure and improved its translational robustness. On the contrary, hybridizing INDI outer-loop with NMPC seems challenging owing to its non-cascaded nature.}

\begin{figure}
    \centering
    \includegraphics[width=0.5\textwidth]{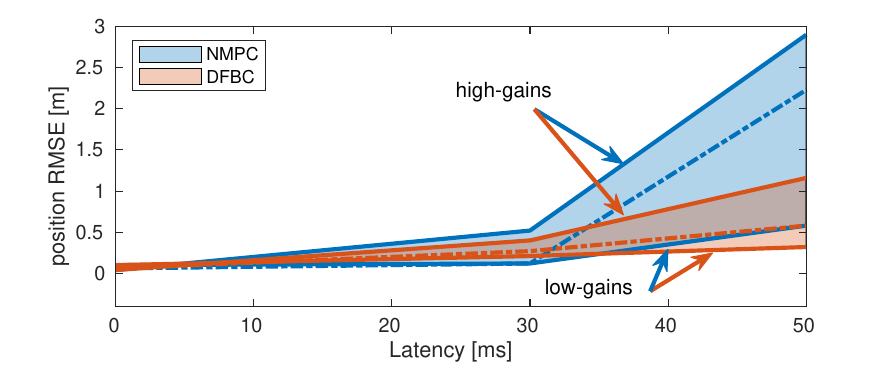}
    \caption{Position RMSE of NMPC and DFBC in the presence of different amounts of estimation latency, with different sets of control gains. Results using gains of previous studies are shown in dash-dot lines. NMPC shows higher sensitivity to the gain selections when estimation latency appears.}
    \label{fig:latency_study}
\end{figure}

The two methods also perform differently in the presence of system latency.
While NMPC slightly outperforms DFBC when the system latency is lower than 30~ms, as the latency grows, NMPC shows a much higher crash rate (\add{68.0}\%) than DFBC (\add{6.7}\%). 
As Fig.~\ref{fig:latency_study} shows, we repeat the tests under 30~ms and 50~ms latency with a high-gain setup and a low-gain setup. In both setups, these gains are tuned such that both controllers have identical average position RMSE in the ideal condition.
As is expected, reducing gains will alleviate the effect of estimation latency. 
Interestingly, we observe that NMPC is more sensitive to the gain selections when system latency is larger than 30~\si{ms}.

\subsection{\add{Effect of Controller Parameters}}
Controller performance is dominated by the parameters or gains. However, finding optimal controller gains can be tedious and highly dependent on the drone parameters. Thus this section performs a sensitivity analysis to provide insights into the effect of controller parameters on the tracking performance. We set the control parameters given in Table~\ref{tab:control_params} as the baseline, and compare the tracking performance with altered parameters with respect to the baseline.

\subsubsection{Nonlinear MPC}
\begin{figure}
    \centering
    \includegraphics[width=0.5\textwidth]{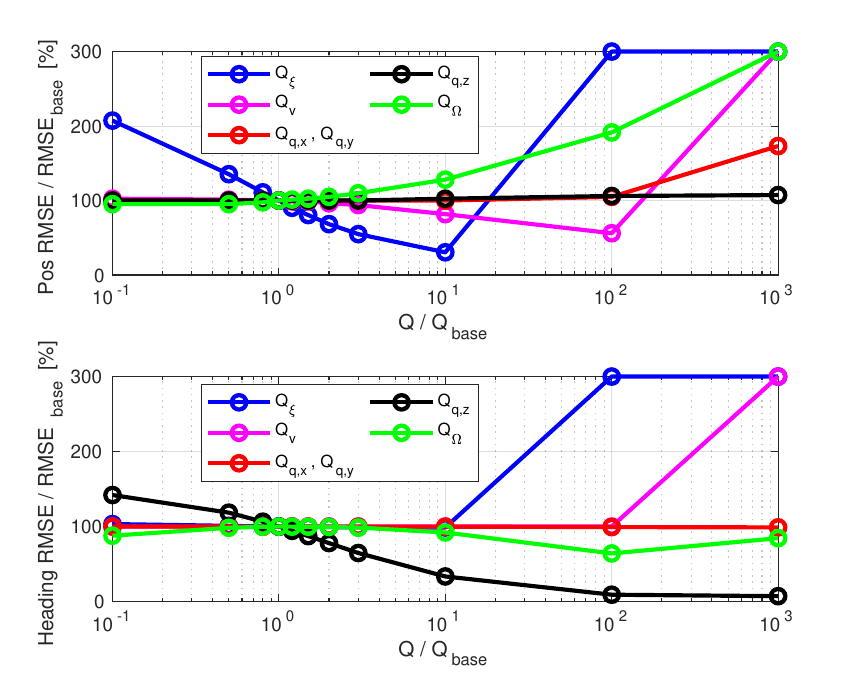}
    \caption{Position (top) and heading (bottom) tracking RMSE of the NMPC method in different control parameters. The vertical axis represents the ratio between RMSE and that from a baseline parameter setup. The horizontal axis represents the ratio between the control parameter and the baseline parameter. The RMSE of those crashed flights are clamped at 300\% of RMSE$_\mathrm{base}$}
    \label{fig:Gain_analysis_NMPC}
\end{figure}

Fig.~\ref{fig:Gain_analysis_NMPC} shows the effect of changing elements of the weight matrix $\boldsymbol{Q}$. The position tracking performance is highly correlated with $\boldsymbol{Q}_{\xi}$ and $\boldsymbol{Q}_{v}$. Among all the parameters, the position weight $\boldsymbol{Q}_{\xi}$ plays the most important role. A larger penalty on the position error results in a smaller tracking error. However, as $\boldsymbol{Q}_{\xi}$ continues growing, the NMPC starts to destabilize the drone. We believe that this is due to the presence of actuator dynamics in the simulation, which introduces system delay and causes instability with high gain controllers. Note that in real-world experiments, system latency can be larger due to state estimation algorithms and signal transmissions, which further reduces the permitted maximum $\boldsymbol{Q}_{\xi}$. Another observation is that increasing $\boldsymbol{Q}_{\Omega}$ also adversarially affects the position tracking performance as it dilutes the power of position weights. Regarding heading control, as is expected, increasing $\boldsymbol{Q}_{q,z}$ can improve the heading tracking performance while causing little effect on position tracking.
\subsubsection{Differential Flatness Based Control}
The controller gains of DFBC consists of position, attitude gains and weight matrix for control allocation. The gains are selected to make the position error $\boldsymbol{\xi}_e = \boldsymbol{\xi} - \boldsymbol{\xi}_r$ a second-order system:
\begin{equation}
    \ddot{\boldsymbol{\xi}}_e + 2\zeta_{\xi}\omega_{n,\xi}\dot{\boldsymbol{\xi}}_e + \omega_{n,\xi}^2\boldsymbol{\xi}_e = 0
    \label{eq:closed_loop_xi_e}
\end{equation}
where $\zeta_\xi$ and $\omega_{n,\xi}$ are damping ratio and natural frequency, respectively. Then substituting (\ref{eq:linear_control1}) into (\ref{eq:closed_loop_xi_e}) and replacing $\ddot{\boldsymbol{\xi}}_d$ with $\ddot{\boldsymbol{\xi}}$ (assuming a perfect inner-loop attitude control), we can calculate the position control gains as follows:
\begin{equation}
    \boldsymbol{K}_\xi = \mathrm{diag(\omega_{n,\xi}^2,~\omega_{n,\xi}^2,~\omega_{n,\xi}^2)},~\boldsymbol{K}_v = 2\zeta_{\xi} \sqrt{\boldsymbol{K}_\xi}.
\end{equation}
Similarly, the attitude gains can be designed as:
\begin{equation}
    k_{q,x} = k_{q,y} = \omega_{n,xy}^2, k_{q,z} = \omega_{n,z}^2,  
\end{equation}
\begin{equation}
    \boldsymbol{K}_{\Omega} = 2\zeta_{q}\mathrm{diag}(\sqrt{k_{q,x}}, \sqrt{k_{q,y}}, \sqrt{k_{q,z}}).
\end{equation}
where $\zeta_{q}$ is the damping ratio, $\omega_{n,xy}$ and $\omega_{n,z}$ are desired natural frequencies of reduced attitude and yaw errors respectively.

\begin{figure}
    \centering
    \includegraphics[width=0.5\textwidth]{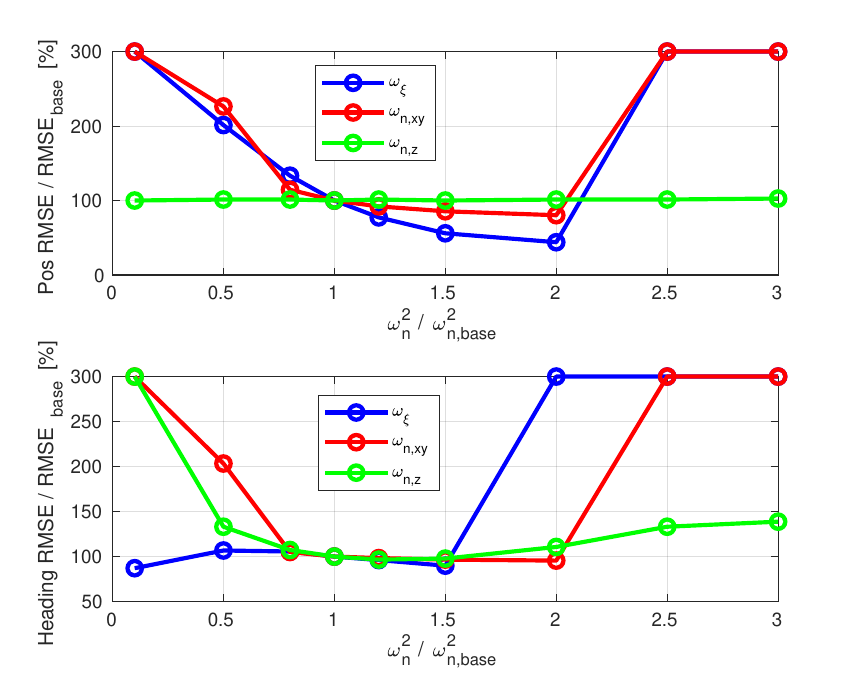}
    \caption{Position (top) and heading (bottom) tracking RMSE of the DFBC method in different control gain settings. The vertical axis represents the ratio between RMSE and the RMSE from a baseline gain setup. The horizontal axis represents the ratio between the controller gains and the baseline gains. RMSE of the crashed flights are clamped at 300\% of RMSE$_\mathrm{base}$}
    \label{fig:Gain_analysis_DFBC}
\end{figure}

Fig.~\ref{fig:Gain_analysis_DFBC} presents the relationship between tracking error and closed-loop natural frequency. Both time constants and tracking errors are normalized by the baseline gains given in Table~\ref{tab:control_params}. A higher frequency indicates a higher control gain, leading to less tracking error. However, high gains are also prone to suffer from system delay and subsequently cause instability.

Weighting matrix $\boldsymbol{W}$ is selected based on tuning in the simulation without sophisticated calculation. It aims at leaving more priority to the tilt control over collective thrust and yaw. An example of inappropriate selection of $\boldsymbol{W}$ is an identical matrix, which makes the QP-based allocation identical to a direct inversion. The performance in terms of crash rate has been compared in Fig.~\ref{fig:crashRates_nonMPC} necessity of using a well-tuned $\boldsymbol{W}$ in tracking dynamically $\textit{infeasible}$ trajectories.

\section{Real-World Experiments}
\label{sec:real_world_experiments}

\begin{table*}[!t]\centering
\caption{Tracking performance of different methods in real-world experiments}\label{tab:tracking_real}
\scriptsize
\begin{tabular}{lcc|ccccccc|cccccccc}\toprule
\multicolumn{3}{c|}{References} &\multicolumn{7}{c|}{Position Tracking RMSE [m]}&\multicolumn{7}{c}{Heading Tracking RMSE [deg]}\\ \midrule
Type &\rotatebox{90}{$V_\mathrm{max}$ [m/s]} &\rotatebox{90}{$a_\mathrm{max}$ [m/s$^2$]}  &\rotatebox{90}{NMPC+PID~\cite{foehn2021time}} &\rotatebox{90}{NMPC}  &\rotatebox{90}{NMPC+INDI} & \rotatebox{90}{\parbox{2cm}{NMPC+INDI\\(w/o drag model)}}  &\rotatebox{90}{DFBC}& \rotatebox{90}{DFBC+INDI} &\rotatebox{90}{\parbox{2cm}{DFBC+INDI\\(w/o drag model)}}  &\rotatebox{90}{NMPC+PID~\cite{foehn2021time}}&\rotatebox{90}{NMPC}  &\rotatebox{90}{NMPC+INDI}& \rotatebox{90}{\parbox{2cm}{NMPC+ INDI \\(w/o drag model)}}   &\rotatebox{90}{DFBC}  & \rotatebox{90}{DFBC + INDI}&\rotatebox{90}{\parbox{2cm}{DFBC+ INDI \\(w/o drag model)}} \\\midrule
Loop A &10.2 &20.7 &0.280 &0.548 &\textbf{0.068} &0.215 &0.897 &0.073 &0.242 &6.801 &11.274 &8.450 &8.972 &73.842 &\textbf{4.411} &9.395 \\
Loop B &13.4 &35.9 &0.369 &0.930 &\textbf{0.117} &0.261 &- &0.125 &0.252 &11.618 &22.193 &\textbf{9.714} &10.478 &- &10.496 &11.489 \\
Slant Loop A &8.2 &12.6 &0.241 &0.458 &\textbf{0.077} &0.156 &0.535 &0.109 &0.163 &6.951 &19.795 &\textbf{5.987} &6.870 &17.755 &7.521 &8.032 \\
Slant Loop B &12.9 &30.8 &0.447 &0.777 &\textbf{0.116} &0.217 &- &0.138 &0.305 &7.729 &18.382 &\textbf{8.282} &8.816 &- &9.171 &10.047 \\
Vertical Loop A &7.7 &14.6 &0.197 &0.206 &\textbf{0.069} &0.172 &0.181 &0.082 &0.127 &1.558 &5.131 &\textbf{1.108} &1.274 &30.901 &1.151 &1.530 \\
Vertical Loop B &11.5 &32.9 &0.482 &0.350 &\textbf{0.126} &0.314 &- &0.177 &0.196 &19.000 &6.019 &2.415 &\textbf{1.900} &- &15.495 &8.632 \\
Oscillate A &10.9 &20.7 &0.327 &0.212 &\textbf{0.058} &0.259 &0.310 &0.164 &0.210 &1.641 &7.369 &\textbf{2.029} &1.960 &50.210 &3.977 &1.911 \\
Oscillate B &14.0 &36.5 &0.375 &0.290 &\textbf{0.125} &0.377 &- &0.129 &0.294 &2.708 &9.228 &\textbf{2.386} &2.902 &- &2.763 &3.821 \\
Hairpin A &7.7 &14.6 &0.185 &0.352 &\textbf{0.024} &0.146 &0.350 &0.049 &0.124 &4.733 &7.100 &\textbf{3.745} &4.255 &38.980 &5.913 &5.873 \\
Hairpin B &11.5 &32.9 &0.390 &0.479 &\textbf{0.099} &0.274 &1.797 &0.117 &0.166 &7.965 &15.748 &\textbf{7.343} &8.017 &68.139 &10.163 &8.004 \\
Lemniscate A &9.5 &13.6 &0.225 &0.324 &\textbf{0.056} &0.184 &0.423 &0.068 &0.216 &3.576 &9.074 &3.689 &4.109 &37.235 &\textbf{2.605} &2.243 \\
Lemniscate B &14.2 &36.0 &0.314 &0.564 &0.130 &0.290 &- &\textbf{0.128} &0.287 &7.335 &15.952 &6.280 &6.659 &- &\textbf{4.986} &5.403 \\
Split-S &14.2 &25.5 &0.292 &0.352 &0.096 &0.262 &- &\textbf{0.087} &0.239 &9.411 &13.253 &6.409 &\textbf{5.259} &- &17.888 &18.929 \\
Racing Track A &11.9 &22.1 &0.280 &0.424 &0.108 &0.212 &0.552 &\textbf{0.102} &0.240 &6.806 &10.434 &6.566 &7.706 &37.731 &\textbf{4.336} &5.390 \\
Racing Track B &16.8 &28.5 &0.369 &0.546 &\textbf{0.130} &0.320 &3.797 &0.152 &0.359 &11.930 &14.672 &9.318 &8.805 &57.810 &\textbf{8.249} &7.719 \\
Racing Track C &19.4 &37.3 &0.712 &0.758 &\textbf{0.238} &0.329 &- &0.259 &0.480 &11.235 &19.936 &11.754 &\textbf{9.876} &- &14.695 &15.498 \\
\midrule
\multicolumn{3}{c|}{Mean} &0.343 &0.473 &\textbf{0.102} &0.249 &0.982 &0.122 &0.244 &7.562 &12.848 &\textbf{5.967} &6.116 &45.845 &7.739 &7.745 \\
\midrule
\multicolumn{3}{c|}{\add{Standard deviation}} &\add{0.130}	&\add{0.207}	&\add{0.067}	&\add{\textbf{0.048}}	&\add{1.161}	&\add{0.164}	&\add{0.051} &\add{4.499}	&\add{5.434}	&\add{\textbf{3.054}}	&\add{3.136}	&\add{18.175}   &\add{4.768}	&\add{4.964} \\
\bottomrule
\end{tabular}
\end{table*}
\begin{table}[!ht]\centering
\caption{Performance in tracking three dynamically infeasible trajectories in real-world experiments. Position and heading RMSE comparisons between NMPC and DFBC with INDI inner-loop.}\label{tab:physically_infeasible}
\scriptsize
\begin{tabular}{m{1.6cm}m{0.35cm}m{0.6cm}|rr|rrr}\toprule
\multirow{2}{*}[-1em]{Reference} &\multirow{2}{*}[-1em]{\makecell{$V_\mathrm{max}$ \\$[$\si{m}$]$}} &\multirow{2}{*}[-1em]{\makecell{$a_\mathrm{max}$\\$[\mathrm{m/s^2}]$}} &\multicolumn{2}{c|}{Position RMSE [m]} &\multicolumn{2}{c}{Heading RMSE [deg]} \\\cmidrule{4-7}
& & &\makecell{NMPC\\+INDI} &\makecell{DFBC\\+INDI} &\makecell{NMPC\\+INDI} &\makecell{DFBC\\+INDI} \\ \midrule
Vertical Loop C&12.1 &48.5 &\textbf{0.318} &0.919 &\textbf{2.469} &25.608 \\
Loop C&15.7 &49.3 &\textbf{0.762} &1.829 &\textbf{13.439} &25.876 \\
Lemniscate C&19.0 &54.5 &\textbf{0.570} &1.237 &\textbf{11.680} &53.262 \\
\midrule
\multicolumn{3}{c|}{Average RMSE} & \textbf{0.550} & 1.328 & \textbf{9.196} & 34.915\\\bottomrule 
\end{tabular}
\end{table}

We conduct the experiments on a quadrotor in an instrumented tracking arena. A set of aggressive trajectories are executed to compare the closed-loop tracking performance of NMPC and DFBC in the presence of joint effects such as model mismatch, state estimation error, and system latency. 
The real-world experiments also highlight the contributions of the INDI low-level controller, the aerodynamic force model, to both NMPC and DFBC methods.

\begin{figure*}[t]
    \centering
    \includegraphics[width=0.99\textwidth]{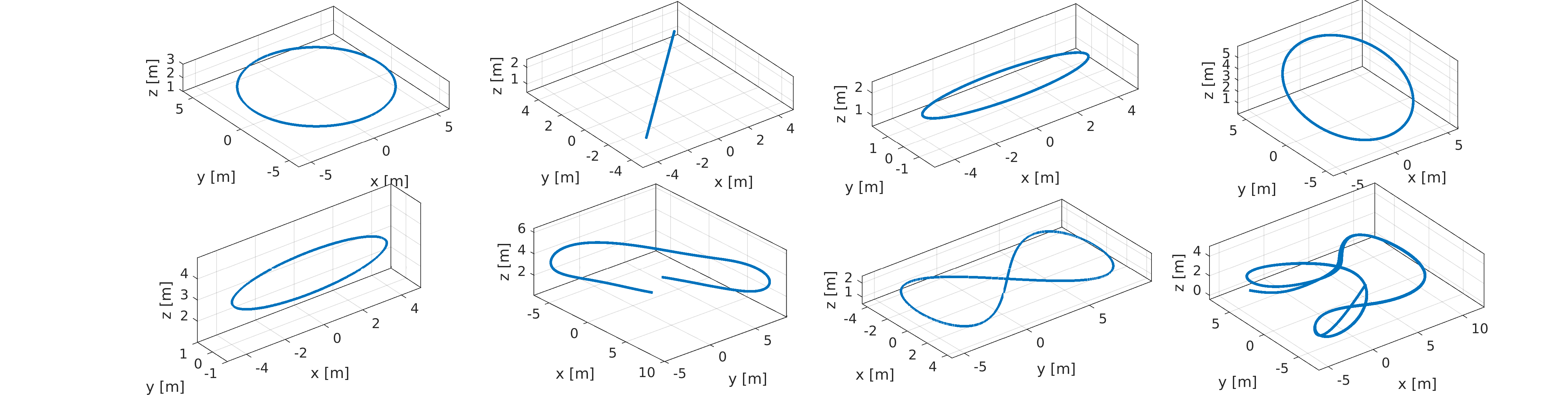}
    \caption{Tested trajectories in real-world experiments. From top-left to bottom right, they are: Loop, Oscillate, Hairpin, Slant-loop, Vertical-loop, Split-S, Lemniscate, Racing-trajectory.}
    \label{fig:ref_trajs}
\end{figure*}

%

Experiments are conducted in tracking different reference trajectories, ranging from simple loop trajectories to FPV drone racing tracks (see Table.~\ref{tab:tracking_real}). Trajectories with the same 3D shape but different velocities and accelerations are also tested. The 3D paths of these trajectories are illustrated in Fig.~\ref{fig:ref_trajs}.

We evaluated NMPC and DFBC, with and without the INDI inner-loop controller.  
NMPC with a PID low-level controller is also tested for comparison, implemented to track time-optimal trajectory in~\cite{foehn2021time}. 
In the following, these methods consider aerodynamic drag models by default, unless further mentioned.
The position tracking RMSE and heading tracking RMSE are listed in Table.~\ref{tab:tracking_real} and Table.~\ref{tab:physically_infeasible}.

\begin{figure*}
    \centering
    \includegraphics[width = 1.0\textwidth]{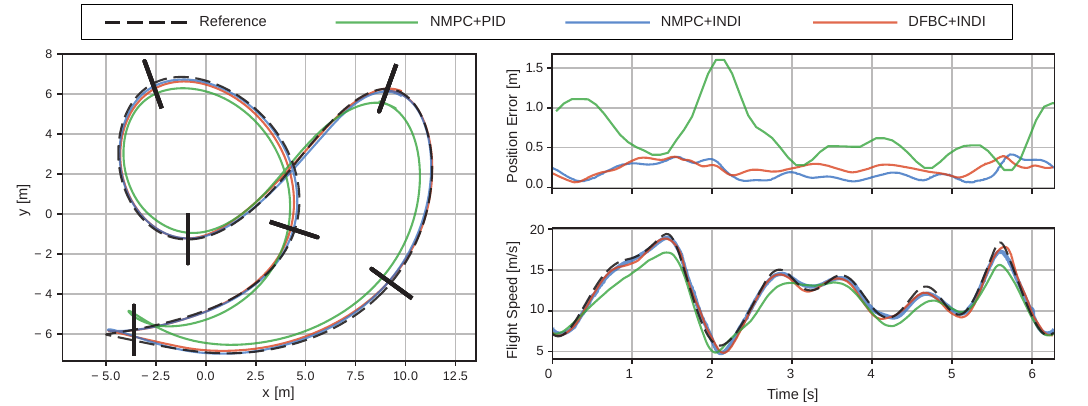}
    \caption{Tracking performance of the \textit{Race Track C} with a maximum speed up to 20~\si{m/s}. NMPC and DFBC with INDI inner-loop show similar tracking performance. Both outperform the state-of-the-art NMPC with a PID inner-loop controller~\cite{foehn2021time}}
    \label{fig:cpc_tracking}
\end{figure*}

\subsection{Contribution of the INDI and Drag Model}
Table~\ref{tab:contribution_INDI_Drag} compares the average RMSE of both methods and shows the effect of the INDI inner-loop controller, and the effect of the aerodynamic drag model.
For NMPC and DFBC, neglecting aerodynamic drag would increase position tracking error by 144\% percent and 122\%, respectively. 

In contrast to the aerodynamic drag model, removing the INDI low-level controller influences more on the tracking accuracy.
For NMPC, we see more than 364\% and 115\% increase in position and heading RMSE if INDI is not used.
For DFBC, removing INDI leads to a more severe consequence: more than 705\% and 492\% of position and heading RMSE increase. 
In fact, DFBC without INDI cannot successfully track some of these trajectories without crashing the drone.
These results indicate that, an adaptive/robust inner-loop controller plays a more important role than the drag model in accurately tracking aggressive trajectories. 

We also make comparisons against the control method used in ~\cite{foehn2021time}, which uses PID as the inner-loop controller of NMPC. 
This comparison is made in tracking a time-optimal trajectory generated off-line using the algorithm proposed in~\cite{foehn2021time}. 
In this task, the quadrotor needs to fly through multiple gates in a predefined order.
Fig.~\ref{fig:cpc_tracking} demonstrates the tracking results, including the 3D path and the position error. 
Clearly, both NMPC and DFBC augmented by INDI significantly outperform this NMPC with PID inner-loop controller. 
Comparative results on other reference trajectories can be found in Table~\ref{tab:tracking_real}.
On average, the proposed NMPC with the INDI approach shows a position RMSE of 0.102~\si{m}, which is 70\% lower than the position RMSE of NMPC with PID low-level controller (0.343~\si{m}).

\begin{table}[!htp]\centering
\caption{Position and heading tracking RMSE of NMPC and DFBC methods. Adding INDI as the inner-loop improves the performance significantly for both methods.}\label{tab:contribution_INDI_Drag}
\scriptsize
\begin{tabular}{l|rl|rl}\toprule
&\multicolumn{2}{c|}{Position RMSE [m]} &\multicolumn{2}{c}{Heading RMSE [deg]} \\\midrule
NMPC+INDI &\textbf{0.102} &$(\uparrow 0\%)$&\textbf{5.967} &$(\uparrow 0\%)$\\
NMPC+INDI (w/o drag model)&0.249 &  $(\uparrow 144\%)$ &6.116 &  $(\uparrow 3\%)$\\
NMPC &0.473 & $(\uparrow 364\%)$ &12.848 &  $(\uparrow 115\%)$\\\midrule
DFBC+INDI &\textbf{0.122} &$(\uparrow 0\%)$ &\textbf{7.739} &$(\uparrow 0\%)$ \\
DFBC+INDI (w/o drag model) &0.271 &$(\uparrow 122\%)$ &7.745 & $(\uparrow 0.07\%)$ \\
DFBC &0.982 &$(\uparrow 705\%)$ &45.845 & $(\uparrow 492\%)$  \\
\bottomrule
\end{tabular}
\end{table}
\subsection{Tracking Dynamically Infeasible Trajectories}
\begin{figure}
    \centering
    \includegraphics[width=0.50\textwidth]{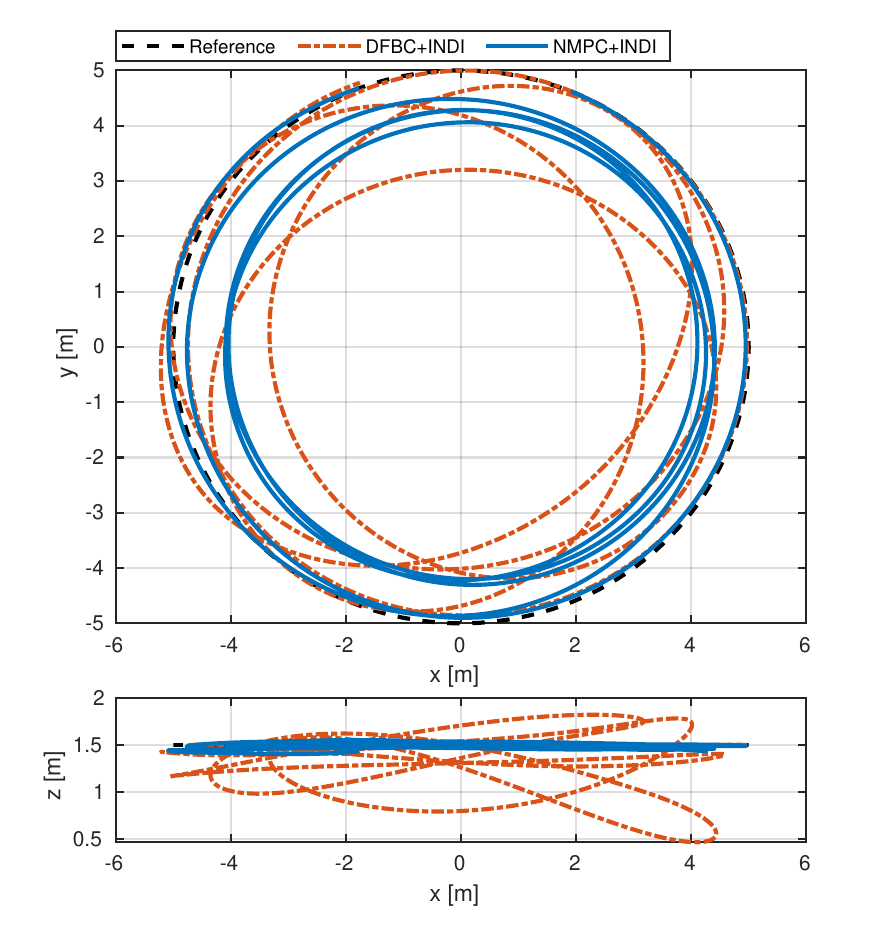}
    \caption{Top and side view of the trajectory tracking a dynamically infeasible trajectory (black-dash) using NMPC (blue-solid) and DFBC method (red-dash-dot). }
    \label{fig:loop16_3d}
\end{figure}

In these real-world experiments, we also track three dynamically infeasible trajectories that exceed the maximum thrust of the tested quadrotor. Comparative results of NMPC and DFBC are given in Table~\ref{tab:physically_infeasible}. Clearly, NMPC shows significantly higher tracking accuracy than DFBC when tracking these dynamically infeasible trajectories, which is \add{consistent} with the simulation results in Sec.~\ref{sec:simulation_experiments}. 

Fig.~\ref{fig:loop16_3d} presents the 3D path tracking the \textit{Loop C} reference trajectory. Tracking such a reference loop trajectory requires a higher collective thrust than the maximum thrust of the tested quadrotor. Hence the quadrotor cannot follow the reference trajectory and both control methods make the quadrotor take the shortcut to fly inside the reference trajectory. While DFBC results in a chaotic trajectory, NMPC makes the drone fly a much more regular loop trajectory with a smaller radius than the reference, thus requiring less collective thrust. The difference on the side-view is more distinctive: NMPC has much higher altitude tracking accuracy than the DFBC method. 

\subsection{Computational Time}
\begin{figure}
    \centering
    \includegraphics[width=0.40\textwidth]{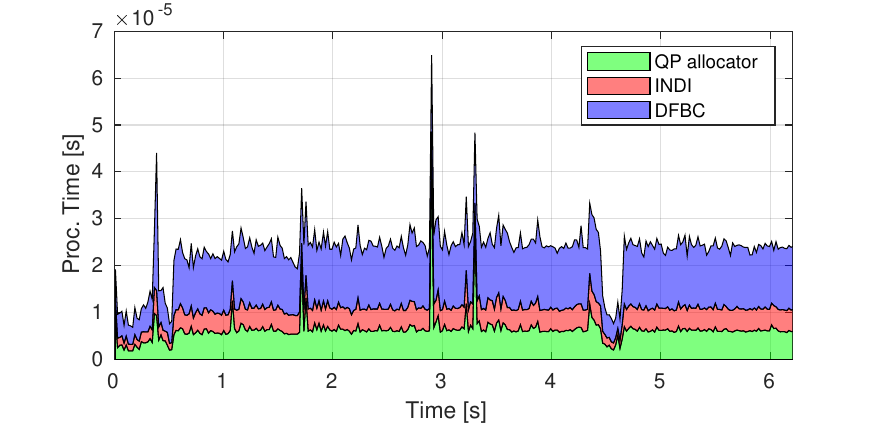}
    \caption{ 
    \add{CPU time of DFBC+INDI while tracking the \textit{Race Track C} trajectory.}}
    \label{fig:proc_time_DFBC+INDI}
\end{figure}

\begin{figure}
    \centering
    \includegraphics[width=0.40\textwidth]{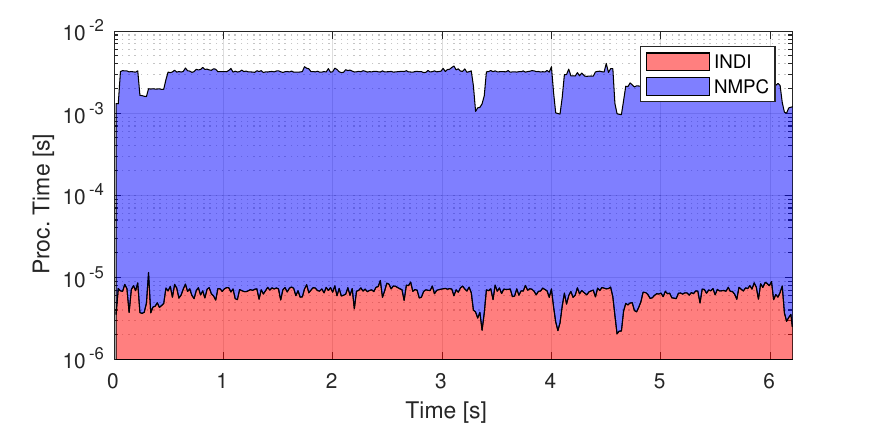}
    \caption{\add{CPU time of NMPC+INDI while tracking the \textit{Race Track C} trajectory. Note that the vertical axis is presented in the log scale.}}
    \label{fig:proc_time_NMPC+INDI}
\end{figure}

\add{
Fig.~\ref{fig:proc_time_DFBC+INDI} and \ref{fig:proc_time_NMPC+INDI} respectively show the CPU time of DFBC and NMPC with INDI as the inner-loop controller while tracking the \textit{Race Track C} trajectory. The computation time of DFBC is generally faster than 0.025~\si{ms}, which is significantly faster compared with NMPC that takes around~3~\si{ms}. In addition, both QP and INDI modules spend less than 0.01~\si{ms} CPU time.
}

\begin{figure}
    \centering
    \includegraphics[width=0.40\textwidth]{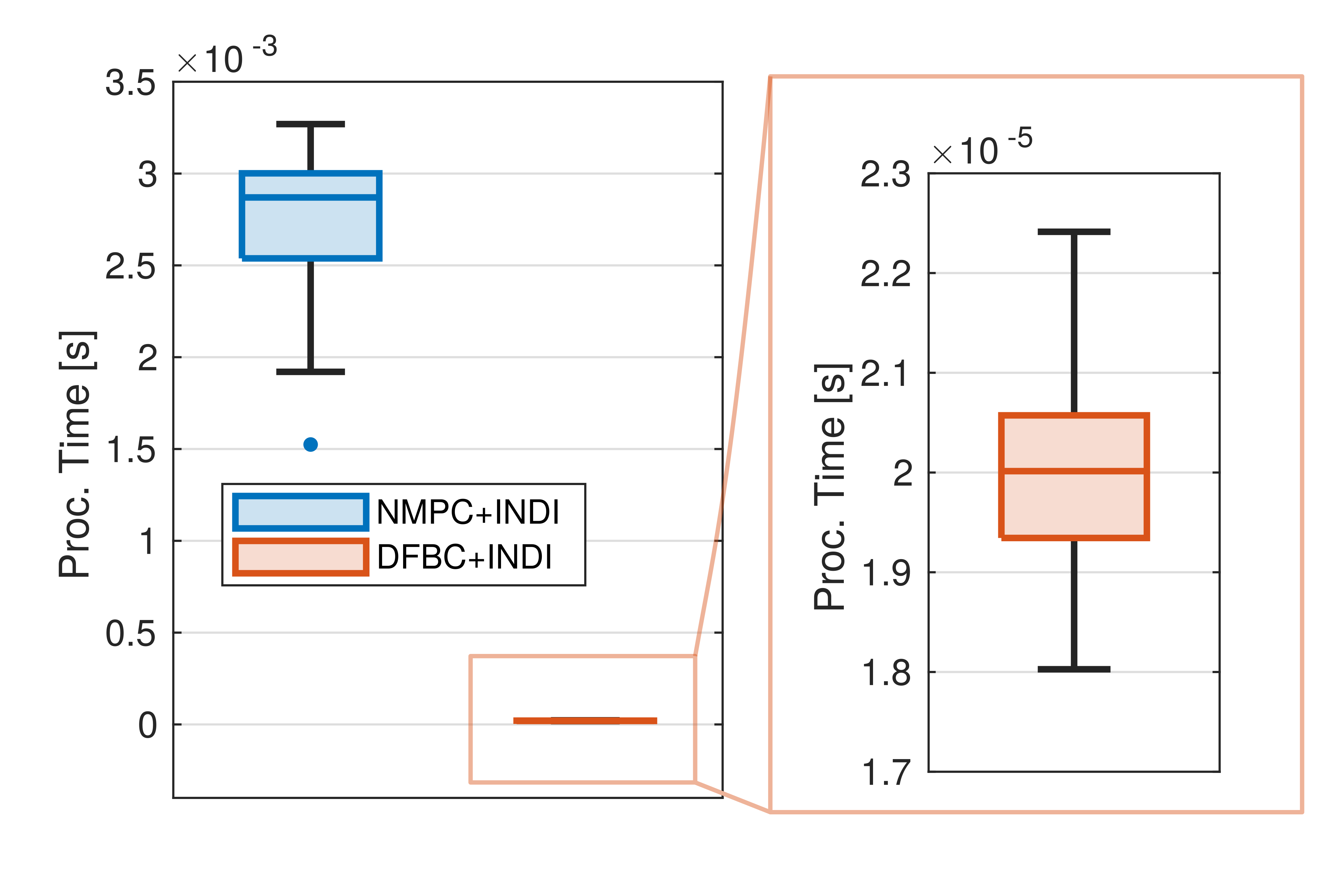}
    \caption{Processing time to generate each control command of NMPC and DFBC methods.}
    \label{fig:timing}
\end{figure}

Fig.~\ref{fig:timing} compares the average computational time of generating each control commands of NMPC and DFBC, with INDI using all the flight data. Each data point in the box plot represents the average computing time of a single flight. As is expected, NMPC requires significantly longer time to generate a single control command (2.7~\si{ms}) compared with DFBC (0.020~\si{ms}). Be that as it may, both methods can run onboard at sufficiently high frequency ($\geq$100~\si{Hz}) to achieve accurate tracking of agile trajectories.
\section{Discussion}\label{sec:discussion}
According to the simulation and real-world flight results, we conclude that both NMPC and DFBC, with an INDI inner-loop controller, can track highly aggressive trajectories with similar accuracy as long as the reference trajectories are dynamically \textit{feasible}. 

The advantage of NMPC appears when tracking dynamically \textit{infeasible} trajectories that violate the rotor thrust constraints. 
Even though DFBC also uses the constrained-quadratic programming for control allocation, it only considers a single reference point; thus, the actions taken are too short-sighted to avoid future violations of these constraints. 
By contrast, NMPC tracks the \textit{infeasible} trajectory using future predictions, including multiple reference points that minimize the tracking error and respect single rotor constraints. 
Such difference helps NMPC outperform DFBC by 48\% and 62\% on position and heading accuracy, respectively, if the reference trajectories are dynamically \textit{infeasible}.

NMPC also demonstrates higher robustness against the rotational model mismatch.
In real-world experiments, we also performed tests without INDI low-level controller for comparisons.
As such, both methods suffered from model uncertainties on the rotational dynamics.
In this condition, NMPC was still able to track all the trajectories, whereas DFBC experienced several crashes and had much higher tracking RMSE against NMPC. 

However, NMPC also has limitations. For example, the biggest disadvantage of NMPC is that it requires significantly higher computational resources. 
On our tested hardware, the average solving time of the nonlinear NMPC is around 2.7 ms, while DFBC needs only 0.020 ms, which is around 100 times faster. 
This renders it impractical to run NMPC on some miniature aerial vehicles with a limited computational budget, such as Crazyflie~\cite{giernacki2017crazyflie}.
By contrast, we can deploy DFBC on light-weighted drones with low-end processors and conduct agile flights at speeds over 20~m/s, as long as the platform has enough thrust-to-weight ratio.   

Another disadvantage of NMPC is that it potentially suffers from numerical convergence issues.
Unlike DFBC, which has proof of stability or convergence of each sub-module, the nonlinear NMPC used in this comparison relies on the numerical convergence of the nonlinear optimization algorithm. 
Unfortunately, rigorous proof of its convergence is still an open question. 
In fact, nonlinear NMPC tends to fail in converging when the current position is too far from the reference, either caused by large external force disturbances or an error in the thrust-to-weight ratio model. 
For example, our robustness study shows a 10\% higher crash rate of NMPC than DFBC when the real mass is 30\% higher than the model. 
In addition, we also observe that NMPC is more prone to fail in converging than DFBC in the presence of large system latency.

In summary, NMPC is not superior to DFBC in all scenarios. If the reference trajectories are dynamically \textit{feasible}, DFBC can achieve similar tracking performance while consuming only 1 percent of the control resource that NMPC requires. 
However, it is difficult to justify the feasibility of a trajectory  near the physical limitations of the platform because of the model uncertainties such as aerodynamic effects. 

Hence, NMPC excels at exploiting the full capability of an autonomous drone in a safer and more efficient manner, due to its advantage in handling the infeasibility of reference trajectories. \add{Future work may further hybridize NMPC with differential flatness property to reduce the NMPC computational cost while retaining its advantage in handling dynamically infeasible trajectories. The reason why NMPC failed in converging also needs further investigation in order to deal with larger amount of model uncertainties.}
\section{Conclusion}
This work systematically compares NMPC and DFBC, two state-of-the-art quadrotor controllers for tracking agile trajectories. 
Simulation and real-world experiments are extensively performed to evaluate the performance of both methods in terms of accuracy, robustness, and computational efficiency. 
We report the advantages and disadvantages of both methods according to the results.
This work also evaluates the effect of INDI inner-loop control on both methods. 
Real-world flight results at up to 20~\si{m/s} demonstrate the necessity of applying an INDI inner-loop on both NMPC and DFBC approaches.
The effect of the high-speed aerodynamic drag model is also evaluated, which is found less influential compared to the inner-loop controller.




\bibliographystyle{ieeetr}
\bibliography{IEEEabrv,references}

\end{document}